\pdfoutput=1

\documentclass[11pt]{article}

\usepackage{acl}

\usepackage{times}
\usepackage{latexsym}
\usepackage{graphicx}
\usepackage{xcolor}

\usepackage[T1]{fontenc}

\usepackage[utf8]{inputenc}

\usepackage{microtype}

\usepackage{multirow}
\usepackage{booktabs}
\usepackage{array} 
\usepackage{makecell} 
\usepackage{graphicx}
\usepackage{soul}
\usepackage{subcaption}
\usepackage{amsmath}
\usepackage[table]{xcolor}
\usepackage{tcolorbox}
\usepackage{makecell}

\definecolor{orange}{rgb}{1,0.5,0}
\definecolor{mdred}{rgb}{0.7,0,0}
\definecolor{mdgreen}{rgb}{0.05,0.6,0.05}
\definecolor{mdblue}{rgb}{0,0,0.7}
\definecolor{dkblue}{rgb}{0,0,0.5}
\definecolor{dkgray}{rgb}{0.3,0.3,0.3}
\definecolor{slate}{rgb}{0.25,0.25,0.4}
\definecolor{gray}{rgb}{0.5,0.5,0.5}
\definecolor{ltgray}{rgb}{0.7,0.7,0.7}
\definecolor{purple}{rgb}{0.7,0,1.0}
\definecolor{lavender}{rgb}{0.65,0.55,1.0}
\definecolor{sangria}{rgb}{0.57, 0.0, 0.04}

\definecolor{darkorange}{RGB}{191, 87, 0}
\definecolor{darkgreen}{RGB}{0, 128, 0}

\usepackage{fontawesome5}
\newcommand{\mainlp}{\faMountain}

\newcommand{\ccs}{\faComments[regular]}
\newcommand{\nrc}{\faCanadianMapleLeaf}

\title{What Media Frames Reveal About Stance: \\A Dataset and Study about Memes in Climate Change Discourse}

\author{Shijia Zhou\kern0pt\textsuperscript{\mainlp\kern0pt}* 
\quad Siyao Peng\kern0pt\textsuperscript{\mainlp\kern0pt}* 
\quad Simon M. Luebke\kern0pt\textsuperscript{\ccs\kern0pt} \\
\large\bf 
Jörg Haßler\kern0pt\textsuperscript{\ccs\kern0pt} 
\quad Mario Haim\kern0pt\textsuperscript{\ccs\kern0pt} 
\quad Saif M. Mohammad\kern0pt\textsuperscript{\nrc\kern0pt} 
\quad Barbara Plank\kern0pt\textsuperscript{\mainlp\kern0pt} \\
\textsuperscript{\mainlp\kern-0pt} MaiNLP \& MCML, LMU Munich, Germany \\ 
\textsuperscript{\ccs\kern0pt} Department of Media and Communication, LMU Munich, Germany\\
\textsuperscript{\nrc\kern0pt} National Research Council Canada, Ottawa, Canada \\
\texttt{\{zhou.shijia, siyao.peng, mario.haim, b.plank\}@lmu.de}}

\newcommand{\corpus}[0]{\textsc{ClimateMemes}}
\newcommand{\convinced}[0]{\textit{convinced}}

\newcommand{\skeptical}[0]{\textit{skeptical}}

\newcommand{\neither}[0]{\textit{neither}}
\newcommand{\real}[0]{\textsc{real}}
\newcommand{\hoax}[0]{\textsc{hoax}}
\newcommand{\cause}[0]{\textsc{cause}}
\newcommand{\causeshort}[0]{\textsc{caus}}
\newcommand{\impact}[0]{\textsc{impact}}
\newcommand{\impactshort}[0]{\textsc{impa}}
\newcommand{\action}[0]{\textsc{action}}

\newcommand{\allocation}[0]{\textsc{allocation}}
\newcommand{\allocationshort}[0]{\textsc{allo}}
\newcommand{\propriety}[0]{\textsc{propriety}}
\newcommand{\proprietyshort}[0]{\textsc{prop}}
\newcommand{\adequacy}[0]{\textsc{adequacy}}
\newcommand{\adequacyshort}[0]{\textsc{adeq}}
\newcommand{\prospect}[0]{\textsc{prospect}}
\newcommand{\prospectshort}[0]{\textsc{pros}}
\newcommand{\ocr}[0]{\begingroup\setlength{\fboxsep}{0pt}\colorbox{magenta!50}{\textcolor{black}{\strut OCR}}\endgroup}
\newcommand{\hum}[0]{\begingroup\setlength{\fboxsep}{0pt}\colorbox{mdgreen!50}{\textcolor{black}{\strut hum}}\endgroup}
\newcommand{\syn}[0]{\begingroup\setlength{\fboxsep}{0pt}\colorbox{orange!50}{\textcolor{black}{\strut syn}}\endgroup}
\newcommand{\fra}[0]{\begingroup\setlength{\fboxsep}{0pt}\colorbox{cyan!50}{\textcolor{black}{\strut frame}}\endgroup}

\begin{document}
\maketitle
\begin{abstract}
Media framing refers to the emphasis on specific aspects of perceived reality to shape how an issue is defined and understood. 
Its primary purpose is to shape public perceptions, often in alignment with the authors' opinions and stances.
However, the interaction between stance and media frame remains largely unexplored.
In this work, we apply an interdisciplinary approach to conceptualize and computationally explore this interaction with internet memes on climate change.
We curate \corpus{}, the first dataset of climate-change memes annotated with both stance and media frames, inspired by research in communication science.
\corpus{} includes 1,184 memes sourced from 47 subreddits, enabling analysis of frame prominence over time and communities, and sheds light on the framing preferences of different stance holders.
We propose two meme understanding tasks: stance detection and media frame detection.
In various input setups, we evaluate two vision language models (VLMs), LLaVA-NeXT and Molmo, and report the corresponding results on their backbone large language models (LLMs).
Human captions consistently enhance performance.
Synthetic captions and human-corrected OCR also help occasionally. 
Our findings highlight that VLMs perform well on stance, but struggle on frames, where LLMs outperform VLMs.
Finally, we draw on concepts from Computational Communication Science to analyze VLMs’ limitations, showing that memes employing specific humor types, personalization, and responsibility cues pose challenges for VLMs in handling nuanced frames and stance expressions on climate change.
\footnote{We make our dataset publicly available at \url{https://github.com/mainlp/ClimateMemes}.}

\end{abstract}

\section{Introduction}

\begin{figure}[t]
    \centering
    \begin{subfigure}[b]{0.45\linewidth}
        \centering
        \includegraphics[height=3.5cm]{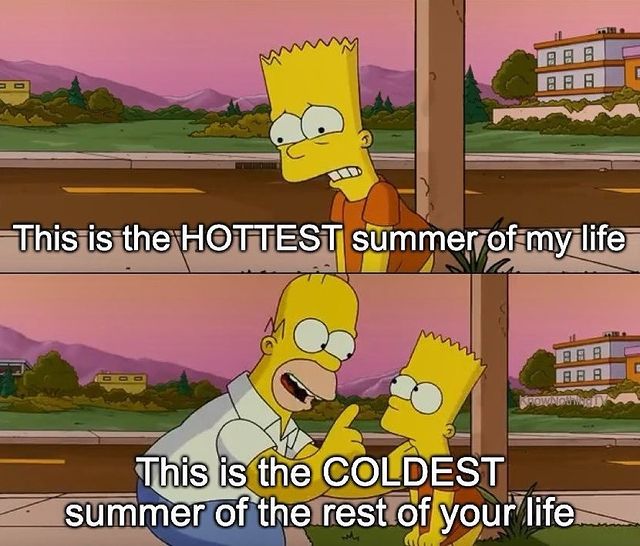}
        \caption{\convinced{} stance with \real{} and \impact{} frames}
        \label{fig:subfigure1}
    \end{subfigure}
    \hfill 
    \begin{subfigure}[b]{0.45\linewidth}
        \centering
        \includegraphics[height=3.5cm]{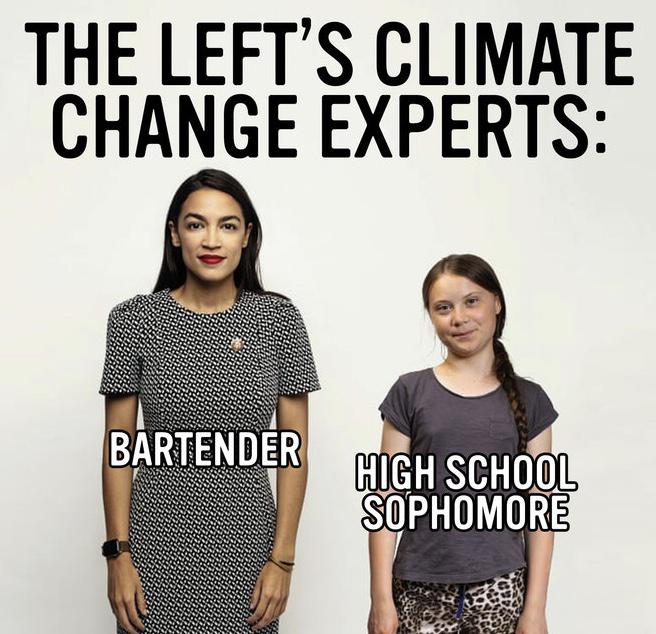}
        \caption{\skeptical{} stance with \hoax{} frame}
        \label{fig:subfigure2}
    \end{subfigure}
    \caption{Two climate change memes conveying opposite stances using different media frames.
    }
    \label{fig:two-memes}
\end{figure} \text

Internet memes are a powerful communication format in online discourse that reflect communities' cultural and social dynamics ~\cite{Davis2016YouCR, Zhang2021ChangingTW}.
Multimodal digital items combine images and texts to convey complex viewpoints in a compact and engaging format~\cite{sharma-etal-2020-semeval, liu-etal-2022-figmemes}.
Through these multimodal expressions, communicators convey their positions towards particular topics, i.e., stances as defined by \citet{mohammad-etal-2016-semeval}.
 
While stance reflects the creator’s opinion toward a target, the specific narrative used to convey a certain stance is shaped by media frames.
Media framing refers to selecting specific aspects of a perceived reality in communication to portray how an issue is defined, how its causes are interpreted, how its moral implications are evaluated, and what potential solutions are considered~\cite{entman1993framing, gidin1980whole}. 
Depending on their stance, creators may gravitate toward different framing strategies~\cite{Snow_Benford_1992}.
However, the interaction between stance and media frames remains under-studied, particularly in their representation through humorous social media content such as memes.
This is especially relevant for debates of global significance, such as climate change.

Memes about climate change (CC) are widespread on social media, including Twitter/X~\cite{ross_internet_2019}. 
For example, Figure~\ref{fig:subfigure1} conveys a \textit{convinced} stance towards CC by using \real{} and \impact{} frames (further detailed in \S\ref{subsec:media_frames_annotation}) to affirm the evidence of global warming and its disheartening consequences.
Conversely, Figure~\ref{fig:subfigure2} conveys a \textit{skeptical} stance using the \hoax{} frame, claiming that CC is not a major issue or even not real, and suggests that politics may distort the CC issue.

In this paper, we analyze stances and media frames in CC memes
by examining the following three research questions (RQs): 

\begin{itemize}
\setlength{\parskip}{0pt} 
\setlength{\topsep}{0pt}  
\item 
\textit{RQ1: How do different media frames shape the visual representation of climate change in memes across varying stances?}
We introduce \corpus{}, a dataset of CC memes, consisting of 1,184 CC memes from 47 subreddits, manually annotated with stance on climate change and the media frames they invoke (§\ref{sec:dataset}) to analyze how memes convey stance through strategic media framing (§\ref{sec:stance_frame}).

\item 
\textit{RQ2: Can state-of-the-art VLMs and LLMs accurately detect stances conveyed by memes and the corresponding media frames?}
We extend \emph{stance detection} from text and propose a new task of multi-label \emph{media frame detection} on CC memes. 
We evaluate two open-source VLMs and their backbone LLMs (§\ref{subsec:learning_setups}) and investigate the effects of few-shot experiments and input modalities on these two tasks (§\ref{subsec:result_main}).
We found that 
while synthetic meme captions cannot yet fully replace human-annotated ones, they still improve the VLMs' performance on both tasks. 
Yet,  LLMs outperform VLMs on frame detection.

\item 
\textit{RQ3: Can taxonomies from communication science provide more insights into stance and media frame detection results?}
We recruit communication science specialists to annotate humor type, person, and responsibility features on 235 test CC memes. 
Our analyses reveal that the performances of VLMs and LLMs degrade markedly on memes that are jokes, about political figures, and about
individual (micro-level) responsibilities (§\ref{sec:commuication}). 
\end{itemize}

\section{Background}

\begin{figure*}
    \centering
    \includegraphics[width=0.98\linewidth]{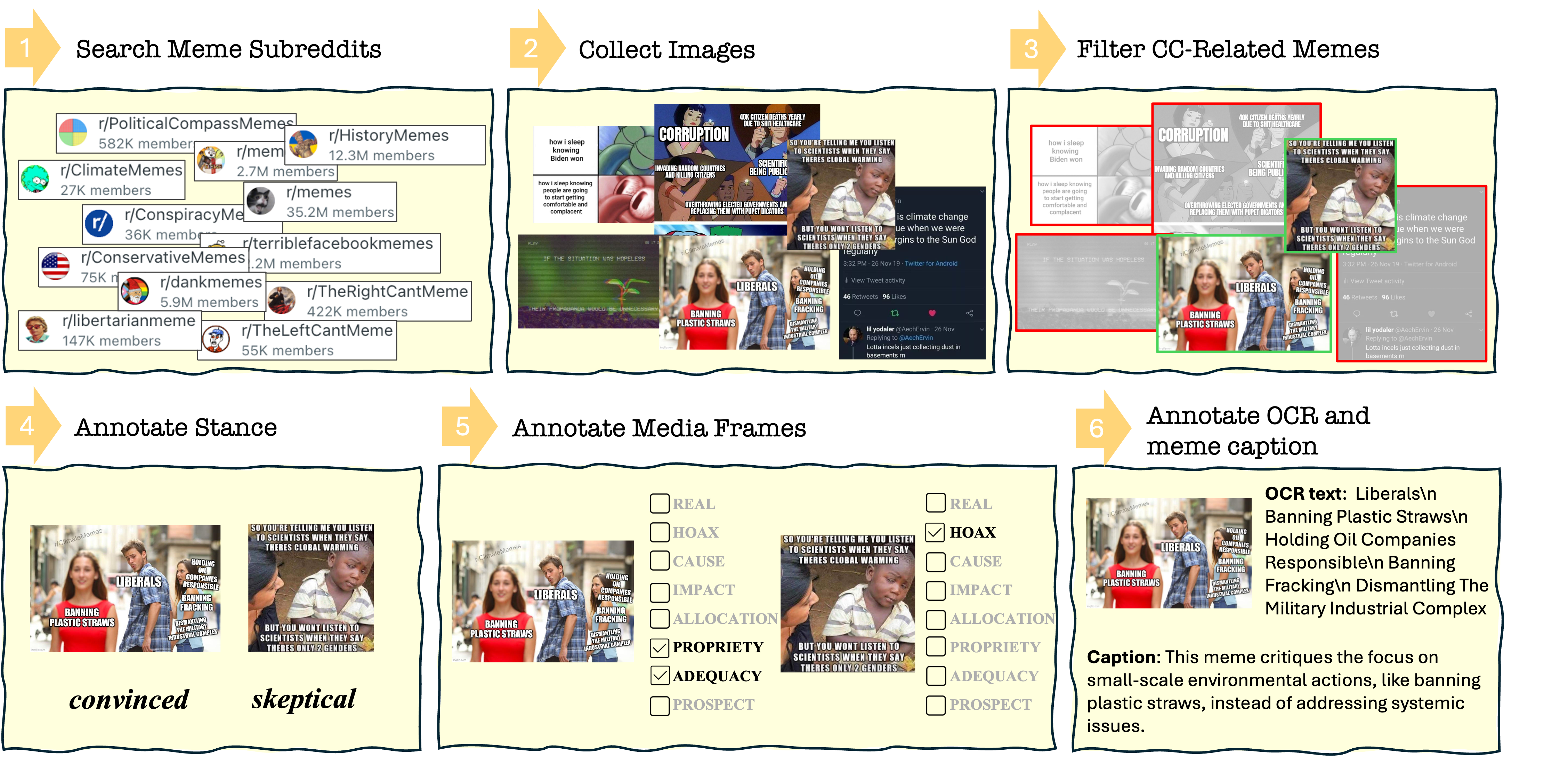}
    \caption{\corpus{}'s pipeline of data collection, filtering, and annotations of stance, media frames, etc.}
    \label{fig:pipeline}
\end{figure*}

\subsection{Memes} \label{subsec:memes}
Internet memes are multimodal and humorous forms of expression that are popular across various digital channels, especially on social media~\cite{shifman_2014}. They often use replicated and modified templates and are circulated among users to convey new, context-specific meanings.
For example, Figure \ref{fig:subfigure1} uses the ``Simpsons so far'' template to highlight the ongoing trend of global warming.
In controversial political arenas, such as the climate discourse, internet memes are seen as an effective tool for capturing attention, allowing users to communicate their stances through impactful imagery and humor~\cite{ross_internet_2019}. 

\citet{nguyen-ng-2024-computational} categorize meme understanding into three main types: \emph{classification}, \emph{interpretation}, and \emph{explanation}. Classification aims to assign labels to memes, such as identifying harmful content ~\cite{kiela_hateful_2020, pramanick-etal-2021-detecting, cao-etal-2022-prompting, ijcai2023p665, chen-etal-2025-adammeme, liu-etal-2025-mind, chakravarthi-etal-2025-findings}, sentiment ~\cite{sharma-etal-2020-semeval, chauhan-etal-2020-one, nguyen-etal-2025-memeqa}, or figurative language~\cite{liu-etal-2022-figmemes, 10.1145/3477495.3532019}. Interpretation tasks focus on understanding and generating insights from memes, such as generating captions or analyzing the metaphor between the image and text componants~\cite{hwang-shwartz-2023-memecap, chen2024xmecap}. Explanation tasks go a step further by generating textual justifications for the labels assigned to memes~\citep{ijcai2023p665}.
In this study, we curate \corpus{} and introduce two meme understanding tasks: stance detection and media frame detection.
We also collect human-corrected OCR and human-written meme captioning as a basis for future tasks.

\subsection{Media Frames} \label{subsec:media_frames}

Strategic media framing refers to the 
selective presentation of information
to influence audience attitudes or evoke specific reactions~\cite{Snow_Benford_1992}. Social and communication science research has relied on framing concepts for analyzing how information is selected and presented in the media. Scholars in the field have created codebooks for manual identification of generic and issue-specific frames in media contexts.

The Media Frames Corpus~\citep{card-etal-2015-media}, focusing on three specific issues: immigration, smoking, and same-sex marriage, brought the methodologies of framing into our NLP community.
Subsequent efforts have expanded this foundation, including proposals for general, issue-independent frame taxonomies~\citep{johnson-etal-2017-leveraging}, computational framing analysis approaches~\citep{mendelsohn-etal-2021-modeling, ali-hassan-2022-survey}, and highlighting the importance of cognitive, linguistic, and communicative aspects beyond topical content in frame detection~\citep{otmakhova-etal-2024-media}.

In the context of climate change, framing has been studied to understand its role in public discourse and media representation~\citep{otmakhova-frermann-2025-narrative}.
\citet{stede2023framing} utilize generic frames, 
which are more abstract and commonly observed across political discussions, to analyze climate change in Nature and Science editorials.
\citet{chen2022convergence} study how frames evolve within public events, emphasizing their divergence and convergence in shaping climate change narratives. \citet{frermann-etal-2023-conflicts} analyze how news articles across the political spectrum frame climate change.
To the best of our knowledge, this paper presents the first dataset of multimodal memes annotated with media frames and analyzes how frames interact with stances.

\section{\corpus{} Dataset} \label{sec:dataset}

This section describes \corpus{}, a dataset of 1,184 CC memes from 47 subreddits annotated with media frames and stances. 
Figure \ref{fig:pipeline} illustrates our data processing pipeline. 
We discuss meme collection and climate filtering (§\ref{subsec:source_selection}), and present guidelines for stance and frame annotations (§\ref{subsec:media_frames_annotation}).
We also provide manual OCR correction and meme caption annotation for future uptake (\S\ref{subsec:ocr_meme_cap}).

\subsection{Source Memes and Climate Filter} \label{subsec:source_selection}

\paragraph{Data Source}
To collect CC memes, we search subreddits with  ``meme'' in their names and filter the topic of posts with the keyword ``climate.''
To obtain diverse climate change perspectives, our collection includes subreddits like \texttt{r/ClimateMemes} (primarily hosting climate activists) and \texttt{r/ConservativeMemes} (reflecting a more skeptical community on CC).

Out of 2,015 initially collected images, 1,184 CC-associated memes from 47 subreddits remained after filtering.
Table~\ref{tab:stance_distribution} shows the top~6 subreddits that contribute to 79.6\% of CC-associated memes (see  Appendix~\ref{sec:subreddits} for a complete list of subreddits).
The table also presents distributions of stance and frame labels (to be detailed in \S\ref{subsec:stance_annotation}).

\begin{table}[t]
\centering
\small
\resizebox{0.49\textwidth}{!}{
\begin{tabular}{lrrrrrl}
\toprule
\textbf{\texttt{r/subreddit}} & \textbf{\#m} 
& \textbf{\textit{conv./skep./nei.}} & \textbf{\#f} &  \textbf{top 3 frames} \\
\hline
ClimateMemes & 591 
&\textbf{94.1} / \enspace3.2\;/\;2.7   &  2.35 
& 
 \textsc{adeq, caus, impa}
 \\
TheRightCantMeme   & 90
&  13.2 /\:\textbf{83.5}\;/\;3.3  &  1.70 &
\textsc{hoax, prop, caus}
\\
dankmemes & 90
& \textbf{82.3} /\:13.3\;/\;4.4   & 1.84 
& \textsc{adeq, impa, real}
\\
memes & 76
& \textbf{92.1} / \enspace1.3 / 6.6  & 1.83 
& \textsc{impa, real, adeq}
\\
meme & 50
& \textbf{80.0} / 16.0 / 4.0  & 1.96 
&  \textsc{adeq, impa, real}
\\
ConservativeMemes & 45 
& 22.2 /\;\textbf{68.9} / 8.9  & 2.02 
&  \textsc{hoax, prop, real}
\\
\hline
Total     & 1,184 
& \textbf{78.0} / 17.2 / 4.8  & 2.11 
&  
\textsc{adeq, impa, hoax }
\\
\bottomrule
\end{tabular}
}
\caption{The number of memes (\#m) in the top 6 frequent subreddits, along with 
percentages of \convinced{}, \skeptical{}, and \neither{} stances,  average number of involved frames (\#f), and top 3 frequently used frames. 
}
\label{tab:stance_distribution}
\end{table}

\paragraph{Filtering CC Memes}

Two master's students in computational linguistics manually annotated all images to ensure a curated collection of CC memes: \textit{climate-associated} and in the format of a \textit{meme}.
Annotators first assess the relevance of these images to climate change, retaining only samples where climate change was a central theme.
They then identify whether a sample qualified as a meme by examining its combination of visual and textual elements, humorous or satirical intent, and relevance to cultural or social contexts.
As Figure~\ref{fig:pipeline} Step 3 shows, tweets containing only text
or lyrical statements paired with images 
are excluded. 

\subsection{Annotation} \label{subsec:stance_annotation}

\paragraph{Stance Annotation} 

The SemEval 2016 shared task \citep{mohammad-etal-2016-semeval} introduced the stance detection task to classify tweets based on whether they are \textit{in favor of}, \textit{against}, or show \textit{neither} stance towards specific targets, one of which was ``Climate Change is a Real Concern.''
We assess the stances of these 1,184 CC memes regarding whether the meme creators are \textit{convinced} that climate change is real, \textit{skeptical}, or \textit{neither} (i.e., cannot tell), following terminologies from social science, particularly~\citet{hoffman2011talking}
(detailed in Appendix \ref{subsec:anno_stance}).

\paragraph{Media Frame Annotation} \label{subsec:media_frames_annotation}

In communication science, media frames are frequently identified to capture different, sometimes conflicting, perspectives on climate change. 
\citet{JANG201511} propose five media frames to examine Twitter conversations on climate change.
These frames include: \real, emphasizing whether the risk of climate change is present; \hoax, questioning the faithfulness of public communication regarding the risk; \cause, attributing the risk significantly to human activities; \impact, highlighting the net negative consequences of the risk; and \action, discussing necessary actions to address the risk.
\citet{ross_internet_2019} apply these five media frames to internet memes and exemplify the contrasting stances of individuals who are \convinced{} of the CC issue and those who remain \skeptical.
Yet, they only present a handful of examples, and a dataset for quantitative analysis and modeling is still missing. 

After adopting these five media frames and through six rounds of annotation revisions, we noticed the overly frequent occurrence of \action{}. 
To provide a more fine-grained analysis of media frames on CC memes, we subdivide the \action\ frame into the following four categories: \allocation, \propriety, \adequacy, and \prospect.

\begin{itemize}
\setlength{\itemsep}{0pt} 
\setlength{\parskip}{0pt} 
\setlength{\topsep}{0pt}  
\item 
\allocation\ captures discussions about the responsibility of certain groups, such as nations, organizations, or even generations, to take action on climate change than others;
\item 
\propriety\ reflects debates on whether current actions are appropriate or effective;
\item  
\adequacy\ highlights critiques regarding whether existing measures are sufficient to address climate risks or more actions are needed;
\item  
\prospect\ explores perceptions of the potential outcomes of positive actions, distinguishing between climate doomists, who view catastrophe as inevitable, and climate risk realists, who believe meaningful prevention is still achievable~\cite{davidson2024climate}.
\end{itemize}

This refinement also allows us to integrate additional dimensions observed in the data, most notably, moral evaluation \citep{entman1993framing, gamson1989media}—without introducing entirely separate frames. We embed moral reasoning within the \allocation\ frame, which inherently concerns fairness, blame, and obligation, thereby capturing moral appeals in both \textit{skeptical} and \textit{convinced} memes while preserving theoretical clarity (see Appendix~\ref{subsec:moral_frame} for guidelines and examples).

\paragraph{Inter-Annotator Agreement}
\label{subsec:iaa}
The first author of this paper annotated stances and media frames on all 1,184 CC memes.
To ensure the consistency of the annotations, we asked one master student in computational linguistics to annotate 200 randomly sampled memes following guidelines in Appendix~\ref{subsec:anno_stance}-\ref{subsec:anno_media_frame}.
We achieved high agreement for stance detection: 0.83 on Cohen's Kappa.
For media frame selection, since we allowed one or more labels per meme, we assess MASI distance and achieve an average score of 0.83.
More, Cohen’s $\kappa$ for all eight frames exceeds 0.7 (see Appendix~\ref{subsec:iaa_appendix}).

\subsection{OCR and Meme Caption}
\label{subsec:ocr_meme_cap}
\corpus{} includes two supplementary annotations: OCR correction and meme caption, as in Figure~\ref{fig:pipeline} Step 6.
For each meme, we extract the embedded text via EasyOCR\footnote{https://github.com/JaidedAI/EasyOCR} and ask the two master students to correct any OCR errors manually. 
We follow \citet{hwang-shwartz-2023-memecap} and ask the annotators to write a concise caption describing the message that the meme conveys.
We further investigate in \S\ref{sec:experimental_setup} whether added explicit textual information helps stance and frame detection. 

\begin{figure}[t]
    \centering
    \includegraphics[width=0.98\linewidth]{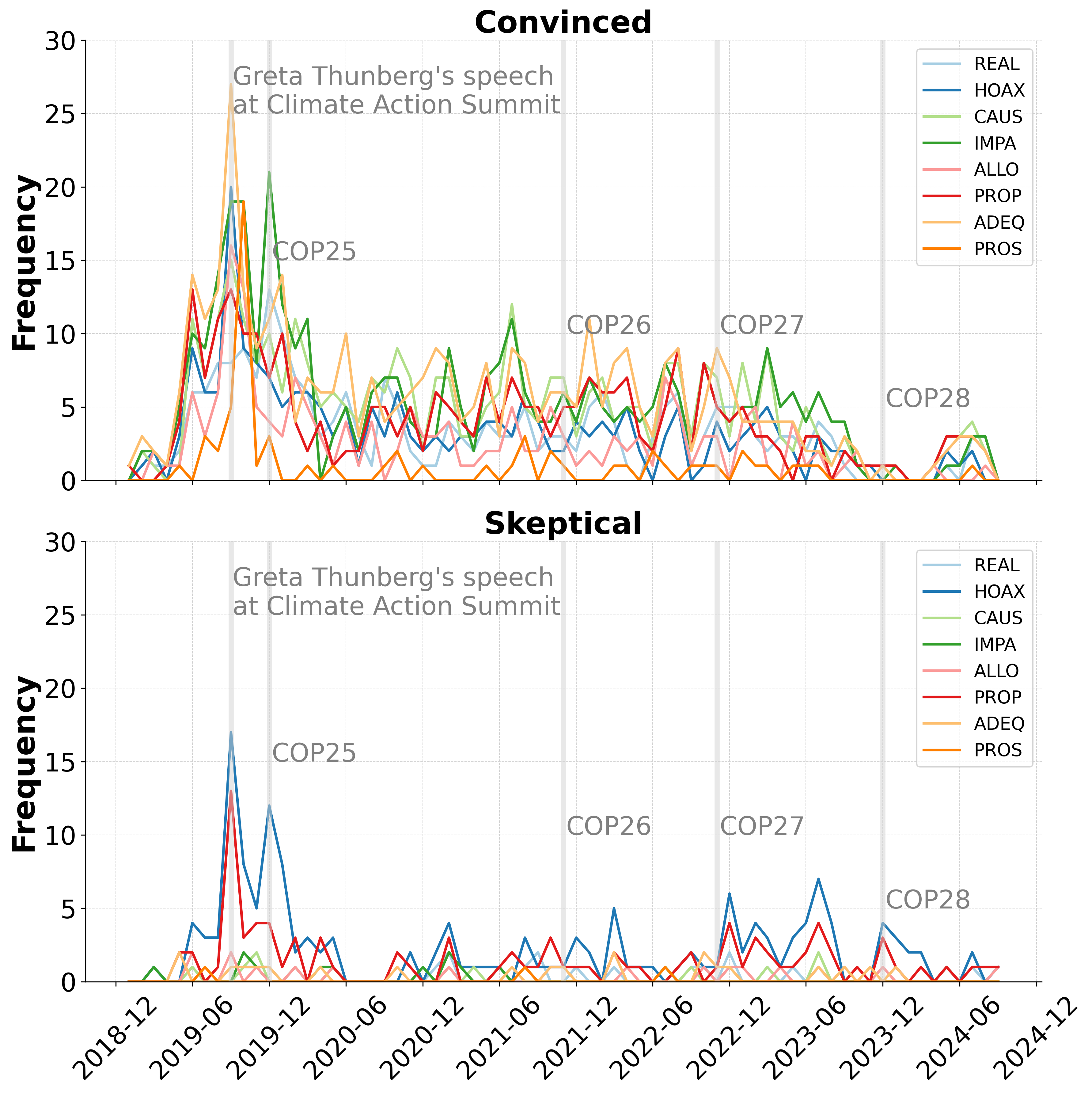}
    \caption{Monthly frequencies of media frames used in \convinced{} versus \skeptical{} memes.}
    \label{fig:shift}
\end{figure}

\section{What Do Media Frames Reveal About Stance?}
\label{sec:stance_frame}

This section analyzes the interactions between stances and media frames in CC memes, including: \corpus{} statistics (\S\ref{subsec:statistics}), frequently used media frames for \convinced{} and \skeptical{} memes (\S\ref{subsec:frame_preference}), concurrences of frames (\S\ref{subsec:frame_concurrence}), and whether specific frames signal a meme’s stance (\S\ref{subsec:frame_signal}).

\subsection{\corpus{} Statistics} \label{subsec:statistics}

Table~\ref{tab:stance_distribution} presents the number of memes in the top 6 frequent subreddits, along with their average number of frames and distribution of \convinced, \skeptical, and \neither{} stances. 
About half of the 1,184 CC-associated memes are sourced from \texttt{r/ClimateMemes}, a community of climate activists.
94.1\% memes from \texttt{r/ClimateMemes} exhibit a \convinced{} stance, with the most frequently occurring frames being \adequacy{}, \cause{}, and \impact{}.
These frames discuss human activities as primary drivers of climate change, enumerate negative consequences, and call for more actions.

\texttt{r/TheRightCantMeme}, \texttt{r/dankmemes} each account for about 8\% of the total memes, ranking second in tie.
83.5\% of the memes from \texttt{r/TheRightCantMeme} demonstrate a \skeptical{} stance, with the predominant frames being \hoax{}, \propriety{}, and \cause{}.
These frames reflect skepticism toward the truthfulness of the CC communications, the effectiveness of current actions, and the denial of human activity as the primary cause.
In contrast, 82.3\% of \texttt{r/dankmemes}  memes exhibit a \convinced{} stance, with \real{} being a common frame, highlighting that CC is indeed happening.

Despite continuous efforts to upsample \skeptical{} memes and subreddits, \corpus{} exhibits an imbalance where 78.0\% memes are \convinced{} and 17.2\% are \skeptical, most frequently employing \adequacy{} and \hoax{} frames, respectively (see Appendix~\ref{sec:frame_distribution} for a detailed frame distribution).

\subsection{Frame Preference} \label{subsec:frame_preference}

This subsection analyzes the framing preference of \convinced{} versus \skeptical{} stances over time.
The publication time of 1,184 \corpus{}  memes spans eight years from March 
2016 to September 
2024.
Figure~\ref{fig:shift} plots the monthly frequency of each frame separately for memes with \convinced{} and \skeptical{}  stances from December 2018 to December 2024.\footnote{CC memes were quite rare before 2019.}
Two peaks occurred in September and December 2019, corresponding to Greta Thunberg’s speech at the United Nations Climate Summit and the COP25, for both \convinced{} and \skeptical{} memes.
Interestingly, in \convinced{} memes, the frequency of nearly all frames is significantly higher during these months, while in \skeptical{}, only the \hoax{} and \propriety{} show an increase.

\begin{figure}[t]
    \centering
    \includegraphics[width=0.98\linewidth]{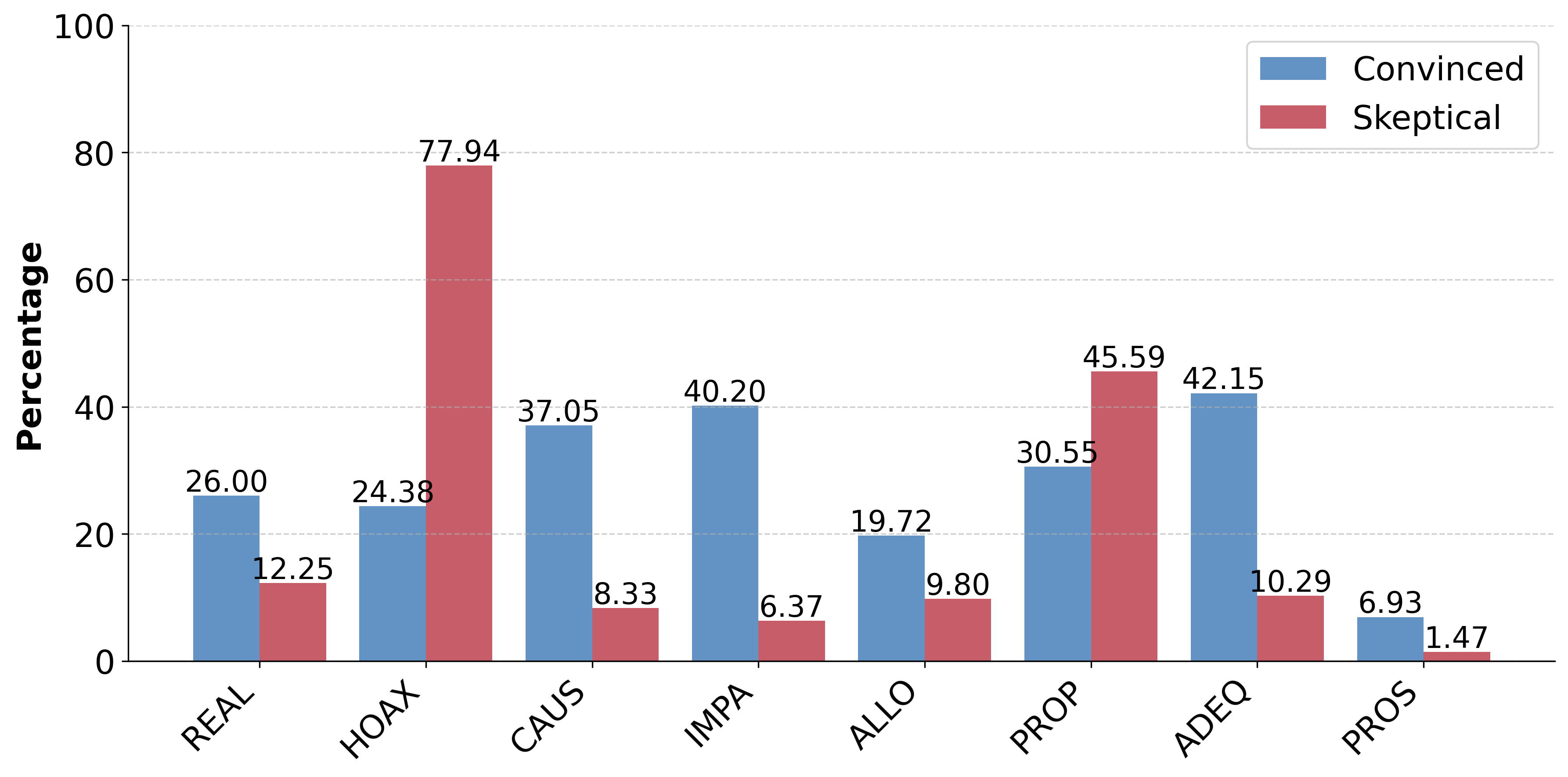}
    \caption{Frame preference of \convinced{} and \skeptical{} memes.}
    \label{fig:frame_preference}
\end{figure}

Figure~\ref{fig:frame_preference} shows the probability of particular frames being involved in memes with \convinced{} and \skeptical{} stances. Among \skeptical{} memes, 77.94\% involve \hoax{}, followed by \propriety{} at 45.59\%.
Other frames appear in less than 15\% of memes.
In contrast, frames in \convinced{} memes are more diverse, with \adequacy{}, \impact{}, and \cause{} being the most common, appearing in 42.1\%, 40.20\%, and 37.05\% of memes.
Other frames, except for \prospect{}, appear in 20\%-30\% of memes.

\subsection{Frame Concurrence}\label{subsec:frame_concurrence}
\begin{figure}[t]
    \centering
    \begin{subfigure}[b]{0.49\linewidth}
        \centering
        \includegraphics[width=\linewidth]{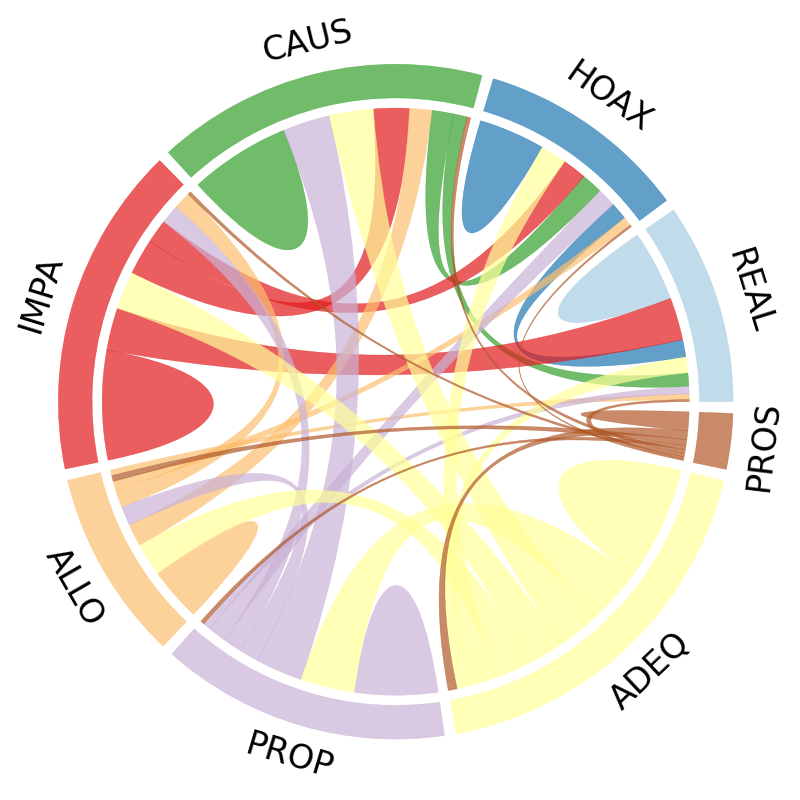}
        \caption{convinced}
        \label{fig:chord_subfigure1}
    \end{subfigure}
    \hfill 
    \begin{subfigure}[b]{0.49\linewidth}
        \centering
        \includegraphics[width=\linewidth]{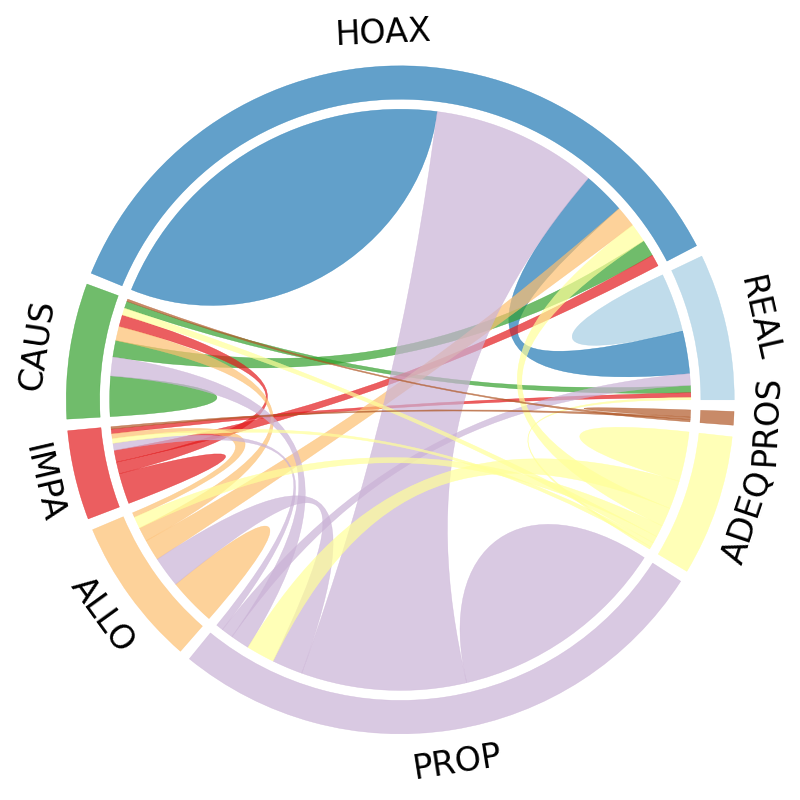}
        \caption{skeptical}
        \label{fig:chord_subfigure2}
    \end{subfigure}
    \caption{Concurrence of media frames in \convinced{} and \skeptical{} memes.}
    \label{fig:chord}
\end{figure}

Since each meme can use multiple frames (2.11 frames/meme, cf. Table~\ref{tab:stance_distribution}), Figure~\ref{fig:chord} investigates the concurrence of frames in \convinced{} and \skeptical{} memes.
For \skeptical{}, the concurrence of \hoax{} and \propriety{} frames is notably more potent than others. 
Rather, frame concurrences in \convinced{} memes are more balanced across diverse combinations, similar to observations in Figure~\ref{fig:frame_preference}. 
Moreover, we notice that \hoax{} has negative correlations with \cause{}, \impact{}, \adequacy{}, and \prospect{}, i.e., they tend not to co-exist (see Appendix \ref{sec:frame_correlation}).

\subsection{Frame as a signal} \label{subsec:frame_signal}

\begin{figure}[t]
    \centering
    \includegraphics[width=0.98\linewidth]{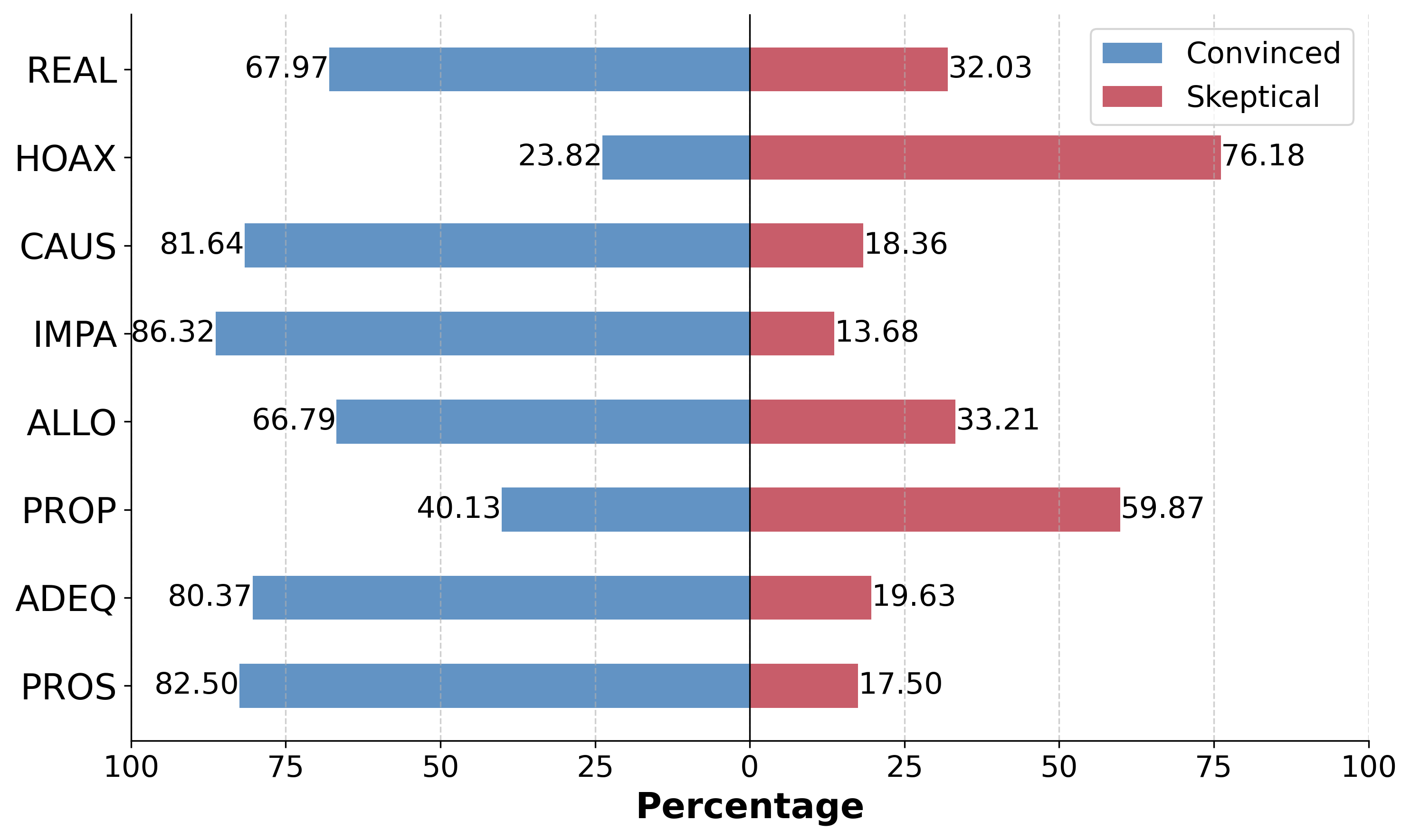}
    \caption{Stance distribution (in percentage) conditioned on frame usage in memes.}
    \label{fig:frame_bias}
\end{figure}

Given that specific frames such as \hoax{} are prevalent in \skeptical{} memes, we examine whether frames serve as a good signal for stance detection. 
Figure~\ref{fig:frame_bias} analyzes the likelihood of a meme being \convinced{} or \skeptical{} when a specific frame is used.
We observe that when \cause{}, \impact{}, \adequacy{}, and \prospect{} appear in a meme, there is $>$80\% probability that the meme holds a \convinced{} stance. \real{} and \allocation{} also appear more frequently in \convinced{} memes. Conversely, \hoax{} implies a 76.18\% probability that the meme is \skeptical{}, followed by \propriety{} (59.87\%).

To sum up, strategic media framing is essential in conveying stances in CC memes. 
Though \hoax{} remains dominant in \skeptical{} memes, framing is more diverse for \convinced{} ones. 

\section{Stance and Media Frame Detection}
\label{sec:experimental_setup}

\textit{To what degree can VLMs detect stance and frames in a meme, and how can we improve their performance?} This section reports various experiments we performed on \corpus{}. 

\subsection{Experimental Setups} \label{subsec:learning_setups}

\paragraph{Models}

We evaluate two open-source VLMs on multimodal memes: LLaVA-v1.6-Mistral-7B (LLaVA, \citealt{Liu_2024_CVPR}) and Molmo-7B-D (Molmo, ~\citealt{deitke2024molmo}), 
both of a \textit{visual encoder→cross-modal connector→LLM} setup. 
To compare, we experiment with text-only inputs on their LLM backbones: Mistral-7B (Mistral, \citealt{jiang2023mistral7b}) and Qwen2-7B (Qwen,~\citealt{, yang2024qwen2technicalreport}). 

\paragraph{Data Partition}
We split \corpus{} into \texttt{train} and \texttt{test} sets with an 8:2 ratio, and all models are evaluated on the \texttt{235} test memes. 

\paragraph{Evaluation Scenarios} 
In addition to zero-shot, we evaluate all models on $n$-shot experiments $n$ ranging from 1 to 4. 
Following \citet{huang-etal-2024-towards}, we leverage relative sample augmentation to select top $n$ similar memes from \texttt{train} for each \texttt{test} meme based on the image and its human-corrected OCR. 
We also explore various input scenarios following \citet{hwang-shwartz-2023-memecap} to examine whether manually-corrected \ocr{} and \hum{}an meme caption (detailed in \S\ref{subsec:ocr_meme_cap}) can improve stance and media frame detection,  as well as \syn{}thetic caption generated by VLMs.
We rotate stance and frame orders in prompts and report the average over permutations~\citep{zheng2023large, wang-etal-2024-answer-c}.
For the backbone LLM baselines, we run experiments on text-only inputs.

\paragraph{Metrics.}
We report accuracy and macro F1 for stance detection, focusing on the latter due to label imbalance.
Since one or more media frames can be assigned to one meme, we binarily classify each frame and report the average over eight frames. 

\begin{table}[t]
\centering
\small

\resizebox{0.98\linewidth}{!}{
\begin{tabular}{ll|rr|rr}
\toprule
\multirow{2}{*}{\textbf{Model}} & \multirow{2}{*}{\textbf{Inputs}} & \multicolumn{2}{c|}{\textbf{Stance}} & \multicolumn{2}{c}{\textbf{Frame}}  \\
 &  & \textbf{F1} & \textbf{Acc.}  & \textbf{F1} & \textbf{Acc.}\\
  \midrule
 baseline & meme  & 29.80 & 80.85 & 43.98 & 73.83\\
  \midrule
 \multirow{6}{*}{LLaVA} & meme  & 39.08 & 77.31 & 45.63 & 51.87\\
  & meme+\ocr{} & 44.06 & 77.31  & 40.72 & 46.36\\
  & meme+\syn{} & 40.01 & 73.95  & 45.78 & 52.45\\
  & meme+\syn{}+\ocr{}   & 41.10 & 76.89  & \textbf{45.87} & \textbf{52.57}\\
  & meme+\hum{} & \textbf{56.68} & \textbf{86.55} & 44.18 & 49.96\\
  & meme+\hum{}+\ocr{} & 53.57 & 83.19 & 44.46 & 50.53\\
    \midrule
 \multirow{6}{*}{Molmo}   & meme  & 28.16 & 47.06 & 52.60 & 60.37\\
  & meme+\ocr{} & 34.70 & 57.56  & 49.68 & 56.98\\
  & meme+\syn{} & 39.25 & 61.76  & 51.02 & 58.37\\
  & meme+\syn{}+\ocr{}  & 38.32  & 65.97  & 47.97 & 54.23\\
  & meme+\hum{} & \textbf{49.53} & \textbf{72.27}  & \textbf{54.24} & \textbf{62.74}\\
  & meme+\hum{}+\ocr{} & 46.52 & 70.17 & 52.46 & 60.40\\

 \midrule
 \midrule

\multirow{5}{*}{Mistral}   & \ocr{} & 37.09 & 51.90 & 54.79 & 61.71\\
 & \syn{} & 36.06 & 58.23 & 53.01 & 59.03\\
 & \syn{}+\ocr{} & 42.71 & 59.66 & 55.20 & 61.78\\
 & \hum{} & \textbf{60.54} & \textbf{79.32} & 58.31 & 64.61\\
 & \hum{}+\ocr{} & 48.96 & 67.65 & \textbf{58.78} & \textbf{65.09} \\
 \midrule

\multirow{5}{*}{Qwen}    & \ocr{}  & 34.06 & 49.16 & 55.45 & 64.02\\
 & \syn{}  & 44.66 & 68.91 & 53.98 & 60.33\\
 & \syn{}+\ocr{}  & 39.08 & 61.34 & 54.24 & 60.88 \\
 & \hum{}  & \textbf{53.28}& \textbf{73.11} & \textbf{58.23} & \textbf{65.86}\\
 & \hum{}+\ocr{}  & 51.66 & 70.17 & 57.51 & 64.98\\
 \bottomrule
\end{tabular}
}
\caption{Performance in accuracy and Macro-F1 on stance and frame detection with 4-shot setup. Backbone LLMs, Mistral and Qwen, only receive text input;
\syn{} = synthetic caption, \hum{} = human caption.
The baseline is calculated using majority vote, detail see Appendix~\ref{sec:majority_vote}.}
\label{tab:main_results_4_shot}
\end{table}

\subsection{Which inputs help stance and frame detection in memes?} \label{subsec:result_main}
Table~\ref{tab:main_results_4_shot} examines how the number of shots and textual inputs influence VLM and LLM performances.

\paragraph{Zero-shot vs. Few-shot}

For both VLMs and their LLM backbones, few-shot setups outperform zero-shot on both tasks, evincing their in-context learning ability (0-4 shots in Appendix~\ref{sec:full_results}).

\paragraph{VLMs vs.\ LLM backbones}
\textit{To what extent can visual inputs benefit VLM performances on meme understanding?}
While LLaVA has an edge over Mistral across various inputs on stance detection (with the exception of meme+syn+OCR and meme+hum for VLMs), both VLMs achieve lower scores on frame detection compared to LLMs.
We hypothesize that VLMs are not pre-trained on meme datasets for frame detection. 
Yet, there already exists textual dataset related to framing \cite{stede-patz-2021-climate, frermann-etal-2023-conflicts}.
It should also be noted that LLMs' winning performances benefit from costly human annotations (OCR corrections\footnote{We observed low-quality OCR; the average Levenshtein edit-distance before and after human correction is 60.75.} and captions) or synthetic captions generated by VLMs. 

\begin{figure}[t]
\centering
\begin{subfigure}[b]{\linewidth}
\centering
\includegraphics[height=5.5cm]{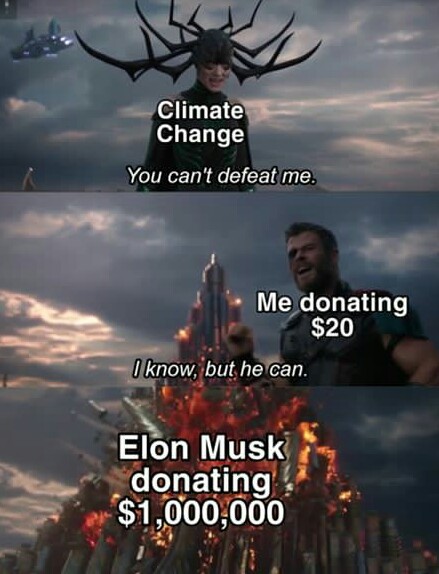}
\caption{\convinced{} stance with \allocation{} and \propriety{} frames}
\label{fig:flame_1}
\end{subfigure}
\hfill 
\begin{subfigure}[b]{\linewidth}
\centering
\includegraphics[height=3cm]{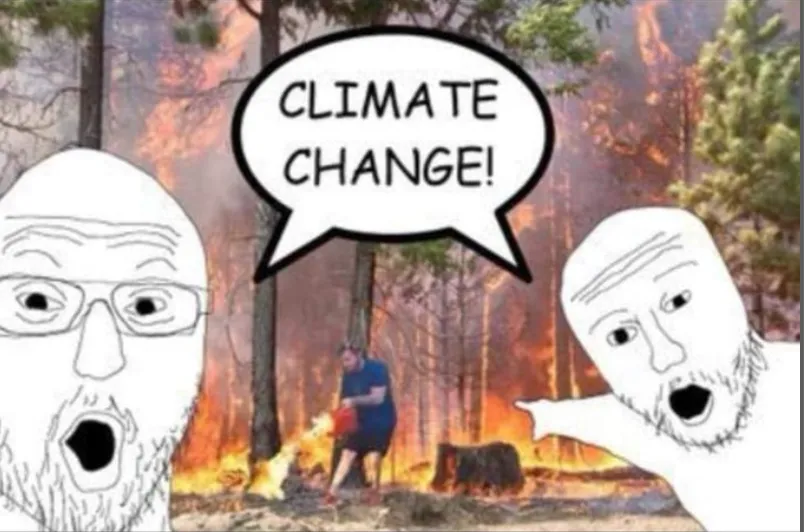}
\caption{\skeptical{} stance with \hoax{} frame}
\label{fig:flame_2}
\end{subfigure}
\caption{Two flame-related memes where LLaVA incorrectly predicts the \textsc{Real} frame.
}
\label{fig:flame-memes}
\end{figure}

Through qualitative analyses, we find that certain visual elements can mislead VLMs.
Figure~\ref{fig:flame-memes} shows two examples where 
LLaVA over-associates flames in image with \real.
The template of Figure~\ref{fig:flame_1} comes from the movie \textit{Thor: Ragnarok}, where the flames on the mountain are used to dramatize a war scene. Figure~\ref{fig:flame_2} satirizes the narrative that attributes all environmental issues -- such as forest fires caused by arson -- to climate change. 
Although flames in our dataset are often associated with the existence of climate change, in these cases they do not depict actual climate change disasters but instead serve as background elements or as examples used to cast doubt on climate change.

 \text

\begin{table}[t]
\centering
\resizebox{\linewidth}{!}{
\begin{tabular}{lr|rr|rr||rr|rr}
\toprule
\multirow{2}{*}{\textbf{Frame}} & \multirow{2}{*}{\textbf{\#M}} & \multicolumn{2}{c|}{\textbf{LLaVA}} & \multicolumn{2}{c||}{\textbf{Molmo}}& \multicolumn{2}{c|}{\textbf{Mistral}} & \multicolumn{2}{c}{\textbf{Qwen}} \\
 & & \textbf{F1} & \textbf{Acc.} & \textbf{F1}  & \textbf{Acc.}  & \textbf{F1}  & \textbf{Acc.}  & \textbf{F1} & \textbf{Acc.}\\
\midrule
\real{} & 44   & \underline{26.84} & \underline{30.69} & \textbf{59.43} & 68.99 & \underline{43.88} & \underline{46.60} & 60.32 & \textbf{72.13}\\
\midrule
\hoax{} & 81 & 46.90 & 51.60 & \textbf{59.43} & \textbf{71.54} & 60.54 & \textbf{71.33} & \textbf{61.46} & 71.01\\
\midrule
\causeshort & 75 & 45.90 & 49.47 & 57.06 & 64.63 & 60.64 & 70.37 & 58.22& 65.85\\
\midrule
\impactshort & 77 & 45.37 & 48.03 & 54.64 & 60.59 & \textbf{61.29} & \textbf{71.33} & \underline{56.03} & \underline{61.54}\\
\midrule
\allocationshort & 49 & 48.05 & 52.55 & 55.22  & 60.00 & 60.17  & 69.95 & 56.99 & 64.36 \\
\midrule
\proprietyshort & 80 & 49.54 & 53.62 & \underline{53.18} & \underline{56.86}  & 56.65 & 64.10 & 57.57 & 64.10\\
\midrule
\adequacyshort & 81 & \textbf{50.50} & \textbf{56.86} & 54.84 & 59.68 & 56.98 & 64.26 & 58.28 & 65.21\\
\midrule
\prospectshort& 13  & 50.69 & 56.86  & 54.74 & 59.63 & 53.58 & 58.99 & 56.99 & 62.71 \\
\midrule
\rowcolor{gray!20}
\textit{Total} & 500
 & 45.47 & 49.96 & 56.07 &  62.74 & 56.72 & 64.61 & 58.23 & 65.86\\
\bottomrule
\end{tabular}
}
\caption{Frame-specific performances with 4-shot meme+\hum{} VLMs and \hum{} LLMs. \textbf{Best} and \underline{worst} scores per model are bolded and underlined.
\#M = number of \texttt{test} memes with the frame label.}
\label{tab:result_frame}
\end{table}

\paragraph{OCR} 
On stance detection, extra OCR input is beneficial for VLMs---though only in setups \emph{without}  human caption. For LLMs, feeding VLM-generated meme captions (syn) functions mostly better than using OCR, especially for Qwen.
Combining OCR with synthetic captions can improve the scores for LLaVA in frame detection but always harms Molmo's performance on both tasks.
Importantly, OCR fails to help VLMs and LLMs further when combined with human captions.
This underlines the importance of high-quality captions, leading to the overall best model for stance. 
Instead, for frames, LLMs outperform VLMs. 
We hypothesize that LLMs better grasp text inputs (especially captions) which aid fine-grained frame detection, while VLMs' performance is lower on frames and benefits less from more explicit texts. 

\paragraph{Human vs.\ Synthetic Caption}
Human meme captions improve performance on both tasks in almost all setups (except for frames with LLaVA).
We leave it to future work to probe how meme captions help models understand stances and frames.

\subsection{Which frames are harder?} \label{subsec:result_frame}

Table~\ref{tab:result_frame} reports per-frame performances of VLMs and LLMs.
Consistent with overall performance, Molmo outperforms LLaVA in predicting all 8 frames.
Molmo scores the highest on \hoax{} and the lowest on \propriety{}.
For LLMs, Qwen outperforms Mistral with overall performance, 
but is not the best in every frame.

\begin{table}[t]
\centering
\resizebox{0.75\linewidth}{!}{
\begin{tabular}{ll|rr}
\toprule
\textbf{Model} & \textbf{Base Input}(+\fra) & $\Delta$\textbf{F1} & $\Delta$\textbf{Acc.}   
\\
\midrule
\multirow{6}{*}{LLaVA} & meme & +0.45 & \cellcolor{gray!20}{-5.04}\\
 &meme+\ocr{}  & +5.26 & \cellcolor{gray!20}{-2.52}\\
 &meme+\hum{}  & +1.06 & 0.00\\
 &meme+\hum{}+\ocr{} & \cellcolor{gray!20}{-7.03} & \cellcolor{gray!20}{-15.54}\\
 &meme+\syn{} & +2.27 & +2.52\\
 &meme+\syn{}+\ocr{} & \cellcolor{gray!20}{-4.74} & \cellcolor{gray!20}{-16.39}\\
\midrule
\multirow{6}{*}{Molmo} & meme & +2.81 & +1.26\\
 &meme+\ocr{}  & +3.33 & +7.15 \\
 &meme+\hum{}  & +4.15 & +5.46\\
 &meme+\hum{}+\ocr{} & \cellcolor{gray!20}{-0.30} & +7.14\\
 &meme+\syn{} & +2.73 & +6.31\\
 &meme+\syn{}+\ocr{} & +3.11 &  +7.56\\
\bottomrule
\end{tabular}
}
\caption{VLM performance changes on stance detection when gold frame labels are added as additional inputs. 
}
\label{tab:stance_results_with_frames}
\end{table}
\subsection{Can frame labels help stance detection?}

Table~\ref{tab:stance_results_with_frames} investigates whether adding gold frame labels helps stance detection on 4-shot VLMs. 
Notably, for LLaVA (better at stance detection), incorporating frame information leads to F1 drops of 7.03 with meme+hum+OCR and 4.74 with meme+syn+OCR, while the other four setups show improvements.
Instead, for Molmo (better at frame detection), adding frame information generally boosts its performance.
Both models further improve their performance in the image and human caption setup.
This suggests that stance and frame detection could benefit from multi-task training, improving performance through shared knowledge.

\section{Meme Understanding through the Lens of Communication Science} \label{sec:commuication}

\begin{table}[t]
\tiny
\centering
\resizebox{\linewidth}{!}{
\begin{tabular}{llr|r@{\hskip 1em}r|r@{\hskip 1em}r}
\toprule
\multirow{2}{*}{\textbf{Concept}} & \multirow{2}{*}{\textbf{Label}} & \multirow{2}{*}{\textbf{\#M}} & \multicolumn{2}{c|}{\textbf{Stance}} & \multicolumn{2}{c}{\textbf{Frame}} \\
& & & \textbf{F1} & \textbf{Acc.}  & \textbf{F1} & \textbf{Acc.}\\
\midrule
\multirow{8}{*}{\begin{tabular}[c]{@{}c@{}}\textit{humor}\\\textit{type}\end{tabular}} 
& irony & 33 & 53.30 & \underline{78.79} & 57.43 & 64.91\\
& compare & 25 & 69.90 & 88.00 & 54.67 & 65.75 \\
& surprise & 21 & 50.42 & 85.71 & 56.82 & 64.58\\
& personif.
& 21 & 82.05 & 95.24 & \textbf{61.39} & \textbf{67.26}\\
& joke & 19  & 52.53 & 84.21 & 50.86 & \underline{58.14} \\
& exagger. & 10 & \textbf{100.00} & \textbf{100.00} & 56.37 & 60.31\\ 
& pun & 5 & \underline{33.33} & 80.00 & \underline{42.86} & 59.69\\
& \cellcolor{gray!20}\textit{Total} 
& \cellcolor{gray!20}134 
& \cellcolor{gray!20}87.42
& \cellcolor{gray!20}63.21 
& \cellcolor{gray!20}62.95 
& \cellcolor{gray!20}54.34\\
\midrule
\multirow{5}{*}{\begin{tabular}[c]{@{}c@{}}\textit{personal}\\\textit{-ization}\end{tabular}}& 
ordinary & 86 & 66.60 & \textbf{88.37} & 54.03 & 62.55\\
& celebrity& 25 & 53.23 & 88.00 & 52.27 & 61.75\\
& politicial & 14 & \textbf{78.79} & 85.71 & \underline{47.93} & \underline{54.80}\\
& NGO & 14 & \underline{37.18} & \underline{50.00}  & \textbf{55.43} & \textbf{66.74}\\
& \cellcolor{gray!20}\textit{Total}
& \cellcolor{gray!20}139 
& \cellcolor{gray!20}78.02 
& \cellcolor{gray!20}58.95 
& \cellcolor{gray!20}61.87 
& \cellcolor{gray!20}54.35\\
\midrule
\multirow{4}{*}{\begin{tabular}[c]{@{}c@{}}\textit{respons}\\\textit{-ibility}\end{tabular}}
& macro & 50 & \textbf{51.06} & 88.00  & \textbf{58.39} & \textbf{64.88}\\
& meso & 37 & \underline{31.43} & \textbf{89.19} & \underline{52.10} & 60.47\\
& micro & 37 & 46.88 & \underline{83.78} & 52.56 & \underline{60.26}\\
& \cellcolor{gray!20}\textit{Total} 
& \cellcolor{gray!20}124 
& \cellcolor{gray!20}86.99 
& \cellcolor{gray!20}43.12 
& \cellcolor{gray!20}61.84 
& \cellcolor{gray!20}54.40 \\
\bottomrule
\end{tabular}
}
\caption{
Llava 4-shot meme+\hum{} results on \texttt{test} subsamples with \textit{humor}, \textit{person}, and \textit{responsibility} labels.
\textbf{Best} and \underline{worst} scores per model are bolded and underlined.
}
\label{tab:case_study}
\end{table}

To understand which aspects challenge models in meme understanding, we apply an interdisciplinary approach, integrating three concepts from communication science research: \textit{humor type}, \textit{personalization}, and \textit{responsibility}.
These concepts are critical in understanding the construction of meaning in affective climate communication on social media and may pose challenges for detection tasks. 
Humor is a key feature of memes as they are usually created with the intention to entertain people ~\cite{taecharungroj2015humour}. Thus, we analyzed different humor types, such as puns, sarcasm, and surprise, which can have varying effects on readers.
Personalization is a common communication strategy in political communication in general, and it simplifies complex political issues by focusing on individual actors instead of groups. 

We recruited two bachelor's students in communication science to annotate \textit{humor type}, \textit{person}, and \textit{responsibility} on 235 \corpus{} \texttt{test} memes.
Our guidelines are adapted from a comprehensive codebook on ``Climate Change and Social Media''\footnote{\url{https://osf.io/3hqdk?view_only=dd6035e7b03542e4a66c2fafa4bf0d7d}} provided by \citet{MaC9892}, allowing
multiple labels per item (detailed definitions in Appendix~\ref{sec:cs_concepts}). 
Table~\ref{tab:case_study} shows the most common labels, number of relevant memes, and LLaVA's subsample performance.

In stance detection, LLaVA performs well on memes with humor types \textit{exaggeration} and \textit{personification}. 
Memes with \textit{pun} and \textit{surprise} are difficult, receiving the lowest F1 score.
Memes with \textit{politicial} and \textit{ordinary} under personalization categories show strong performance, but \textit{NGO} stands out as challenging. For responsibility, memes concerning the \textit{macro}-level are the easiest for the model, while \textit{meso}-level memes are the hardest. Since \textit{micro}/\textit{meso}-level responsibilities address specific individuals (e.g., politicians) or organizations (e.g., companies), we hypothesize that \textit{macro}-level responsibility (e.g., the society) leads to less variation and eases meme understanding.

In frame detection, memes with humor types \textit{pun} and \textit{joke} are the hardest. Under personalization, memes featuring \textit{NGO} or \textit{celebrities} are easier for the model than \textit{political ones}. 
Frame detection on memes attributing responsibility mirrors results on stance detection, with \textit{macro} performing best and \textit{meso}  worst.
In sum, this interdisciplinary annotation using taxonomies from communication science provides insights into aspects that caused difficulties in meme stance and frame detection.
Appendix~\ref{sec:error_analysis} provides additional error analyses.

\section{Conclusion} \label{sec:conclusion}

We introduce \corpus{}, a new benchmark dataset of climate change memes annotated with stance and media frames.
We demonstrate that media frame preferences are strong indicators of stance, with \convinced{} and \skeptical{} stances favoring distinct frames.
We compare VLMs and LLMs and identify challenges in understanding multimodal information.
Our paper also integrates concepts from communication science and reveals which meme aspects challenged the model.

\section*{Limitations}

\paragraph{Potential Bias in the Sample Due to the Platform}
Our dataset is exclusively composed of memes collected from Reddit, which introduces a potential bias. By focusing solely on this platform, we limit the diversity of content that could be found on other platforms like Twitter, Facebook, Instagram, or 4chan. Each platform has its own user base, culture, and way of sharing and discussing memes, which could result in differences in the types of memes that are shared. This platform-specific limitation means that our findings might not be fully representative of meme trends across the internet as a whole. 
Therefore, the trends that we observed in our reddit meme data might not reflect trends beyond the period of investigation and the platform. Reddit has a unique structure, where specific subreddits cater to distinct interests, communities, and ideologies, which could influence the stances and frames adopted in memes. For example, some subreddits may have a higher concentration of memes that are either supportive or skeptical of climate change, while other platforms might exhibit different trends. Memes on Twitter or Instagram could carry different connotations, tones, or styles that might not be as prevalent on Reddit. Thus, the distribution of meme stances and frames could vary significantly across platforms, and a more comprehensive understanding of meme discourse would require analyzing multiple platforms to account for these differences.

\paragraph{Scope and Generalizability}
This study focuses exclusively on climate change, a uniquely salient and persistently active topic with global relevance. Our analysis relies on a theoretical framework explicitly developed and validated for climate change discourse~\cite{JANG201511, ross_internet_2019}, examining context-specific media frames and stances, such as responsibility allocation (ALLOCATION) and the appropriateness of measures (PROPRIETY). As a result, our findings may not be directly generalizable to memes on other topics.
Nevertheless, we believe that similar frames and patterns could apply to other topics, such as COVID-19~\citep{doi:10.1177/0957926520970385}, where the divide between convinced and skeptical stances exists alongside debates about responsibility and policy.

\paragraph{Only a Single Annotator}
We acknowledge the limitation of only a single annotator. 
Despite limited resources, we carefully refined our annotation guidelines through six iterations, totaling about 540 hours of annotation work. 

\paragraph{Monthly Frequency: Sample Size May Be Too Small in Some Months to Derive Conclusions About Temporal Trends}
The monthly frequency of memes in our dataset might not be large enough in certain months to allow for meaningful conclusions about trends or changes over time. If the sample size in a given month is too small, it becomes difficult to accurately detect shifts in meme stances, frames, or topics that may occur over longer periods. This limitation could obscure any subtle trends or variations in the frequency of specific meme types or themes, making it harder to assess how the discourse around a particular subject evolves. For instance, if a meme trend spikes during a specific event but the dataset contains very few memes from that month, it might not reflect the broader public sentiment or provide an accurate representation of the temporal dynamics.

\paragraph{Visual inputs for VLMs} We did not evaluate VLMs without visual input, and using the LLM backbone alone might not be 100\% comparable to running a VLM without image input, because VLMs are fine-tuned on different datasets. 

\section*{Ethics Statement}
All annotations were conducted in accordance with ethical guidelines, ensuring that annotators were not exposed to any psychologically distressing content during the process.
All annotators are paid according to national standards.

\section*{Acknowledgements}
This work is supported by the KLIMA-MEMES project funded by the Bavarian Research Institute for Digital Transformation (bidt), an institute of the Bavarian Academy of Sciences and Humanities. The authors are responsible for the content of this publication.

\bibliography{main}

\appendix


\section{Subreddits in \corpus{}}
\label{sec:subreddits}

\begin{table*}[]
\centering
\tiny
\resizebox{0.98\textwidth}{!}{
\begin{tabular}{lrp{9cm}}
\toprule
\textbf{subreddit}      & \textbf{frequency} & \textbf{description}              \\ \midrule
ClimateMemes            & 591 & The community to share environmental memes of prime quality. We advocate   for climate action through funny captions and satire. Release your inner   Greta, share your dankest decarbonization memes and raise global awareness to   save the planet! Discuss climate strikes, climate change denial and   doomerism, Fridays For Future, facts and news about nature, climate crisis   quotes, ecology, Extinction Rebellion, and the end of the world. \\
TheRightCantMeme        & 91  & Get your fix at left-wing Reddit alternatives: Hexbear and Lemmygrad. Also check out the Discord. \\
dankmemes               & 90  & D A N K   \\
memes & 76  & Memes! A way of describing cultural information being shared. An element of a culture or system of behavior that may be considered to be   passed from one individual to another by nongenetic means, especially   imitation. \\
meme  & 50  & r/meme is a place to share memes. We're fairly liberal but do have a few   rules on what can and cannot be shared.    \\
ConservativeMemes       & 45  & Become a ConservativeMemes subscriber! — Click the JOIN button now, and post   your Conservative Memes later at /r/ConservativeMemes !!!  — If you like political humor, political   memes, politically incorrect memes, or conservative memes, this is the sub   for you! \\
PoliticalCompassMemes   & 39  & Political Compass Memes     \\
terriblefacebookmemes   & 30  & Community for all those terrible memes your uncle posts on facebook               \\
ConspiracyMemes         & 18  & This subreddit is devoted to memes relating to all things   conspiracy.  Things are pretty laid   back around here so all people are welcome. The moderators believe in free speech and try not to moderate comments or posts unless it is absolutely necessary. \\
Memes\_Of\_The\_Dank    & 15  & This is a meme subreddit. That should be obvious by now. Also, it is slowly recovering from bots, and that’s good. \\

libertarianmeme         & 14  & For an end to democracy and tyranny. For more information about our ideology, check out the Mises Institute \\
HistoryMemes            & 11  & A place for history memes about events over 20 years ago.       \\

CommunismMemes          & 9& A place to share memes about communism.       \\
PoliticalMemes          & 7& We're striving for equality here. Not "equality" in the sense   that we'll allow people to post bigoted nonsense or perpetuate a false   equivalency of entities, but "equality" in the sense that we are   all co-inhabitants of this flying rock and need to learn to live together   peacefully.       \\
dank\_meme              & 7& Dank Memes\\
MemeEconomy             & 7&           \\
sciencememes            & 7&           \\
PrequelMemes            & 6& Memes of the Star Wars Prequels.              \\
PresidentialRaceMemes   & 6&           \\
MemeThatNews            & 6& Learn and comment on the news with memes.     \\
AusMemes & 6& The Australia Memes subreddit. Just waiting for a mate. \\
Animemes & 6& A community for anime memes!\\

Marxism\_Memes          & 5& MEMES ARE THE NEW PAMPHLETS JURY NULLIFICATION FOR COMRADE LUIGI! \\
TheLeftCantMeme         & 4& They make a lot of bad Political Memes        \\
MinecraftMemes          & 3& A place to post memes about Minecraft! Our Discord Server can be found in the sidebar below. \\
AnarchyMemeCollective   & 3& A reddit for sharing anarchist memes and for discussing anarchism. If you share your own OC let us know and we   may share if on our other platforms. \\
depression\_memes       & 3& Memes about depression.     \\
Funnymemes              & 3& "Where Laughter Lives: Your Daily Dose of the Funniest Memes!"  \\
marvelmemes             & 2& Welcome to r/marvelmemes: The home of Marvel memes on Reddit!   \\
lotrmemes               & 2& Come on in, have a seat! This subreddit is a warm resting place for all   weary travelers who are fond of Tolkien and his works. We welcome all Tolkien   related content! Grab a pint, a long pipe, and relax. \\
VegMeme  & 2& A place to share animal rights humor, cartoons, image macros etc, because   if you can't have a laugh at the hypocrisy and ignorance of carnists or have   a good-natured laugh at ourselves you will probably become a misanthropic   douchebag.   \\
Jordan\_Peterson\_Memes & 2& Welcome to the official subreddit for Jordan Peterson memes.    \\
animememes              & 2& An anime meme subreddit that's friendly for women, queer people, and   generally marginalized anime fans who want a break from how toxic anime   spaces usually are. Of course, anyone is welcome, but be respectful to the   intention of the space.\\
AvatarMemes             & 2& A subreddit for memes and other humor related to the Avatar franchise.   Jokes based on ATLA, LoK, etc. are welcome.  \\
CoronavirusMemes        & 2& Opening back up due to popular demand, didn’t know people still wanted to   post about the coronavirus. Monkeypoxmemes are allowed. Getting a laugh out of the Coronavirus while we still can, and spreading happiness in a time of distress. \\
SequelMemes             & 1& Memes of the Star Wars Sequels \\
VoluntaristMemes        & 1& Memes for voluntarists and other liberty loving people.         \\
CommunistMemes          & 1& Communism is always the end goal!             \\
SimpsonsMemes           & 1& Memes from The Simpsons!    \\
MemePiece               & 1& The best place to find One Piece memes! We celebrate the comedic and   casual side of the series One Piece. Casual or low effort content, normally   removed from r/OnePiece, is likely welcome! \\
CrusadeMemes            & 1& DEUS VULT \\
MemeReserve             & 1& The Doomsday Global Meme Vault is a fail-safe meme storage sub, built to stand the test of time — and the challenge of natural or economical collapse. Only for the best memes! \\
GameOfThronesMemes      & 1& This subreddit is currently closed. Please check out r/aSongofMemesAndRage for memes based off GOT, ASOIAF, etc.    \\
IncrediblesMemes        & 1& It's showtime               \\
memesITA & 1& Pizza, pasta \& memes.      \\
AnimeMeme               & 1& AnimeMeme for anime memes.  \\
YouBelongWithMemes      & 1& The official meme subreddit for r/TaylorSwift     \\
\bottomrule
\end{tabular}
}
\caption{\corpus{}'s 47 subreddits with their descriptions and meme frequency.}
\label{tab:subreddit_descroption}
\end{table*}

Table~\ref{tab:subreddit_descroption} details the public descriptions and meme frequencies of 47 subreddits in \corpus{}.

\section{Annotation Guidelines}
\label{sec:anno_guidelines}

\subsection{Filtering Climate Change Memes}
\label{subsec:anno_filter_cc_memes}

\paragraph{Is this image associated with the topic of climate change?}
Often images that discuss terms such as “climate change,” “global warming,” “greenhouse gas,” “carbon emission,” “fossil fuel,” “ozone,” “air pollution,” “carbon dioxide emissions,” “deforestation,” “industrial pollution,” “rising sea levels,” “extreme weather,” “melting glaciers,” “ocean acidification,” “biodiversity loss,” “ecosystem disruption,” “carbon capture,” “carbon storage,” “soil carbon,” “renewable energy,” “sustainable practices,” “Paris Agreement,” “Kyoto Protocol,” “carbon tax,” “emissions trading schemes,” “green technology,” “sustainable technology,” and “environmental change” are associated with climate change.

Additionally, if the meme features a well-known environmentalist or a political leader who has made statements related to climate change and environmental protection, it should also be considered as "associated with climate change." If you encounter an unfamiliar person, please use Google to search and confirm.

\paragraph{Is this image a meme? Is it a cartoon?}
Memes are created by taking an existing widespread image and attaching new meaning to it by adding text within the image. 
A political cartoon, also known as an editorial cartoon, uses caricatures and satire to express an artist's opinion on current events, often critiquing political leaders, social issues, or corruption through humor and exaggeration. A cartoon style often features exaggerated characters and simplified forms, and the text is usually in hand-drawn fonts that match the casual, expressive tone of the illustration.
Both memes and political cartoons are considered memes in this study. 

\subsection{Stance Annotation}
\label{subsec:anno_stance}

\paragraph{What is the stance of this CC meme?}
We annotate the stances of CC memes into the following three categories: \convinced, \skeptical{} and \neither.

\begin{itemize}
\setlength{\itemsep}{0pt} 
\setlength{\parskip}{0pt} 
\setlength{\topsep}{0pt}  
\item \convinced{}: Accepts environmental risks, supports regulation of harmful activities, and reflects egalitarian and communitarian values.
\item \skeptical{}: Downplays or denies environmental risks, opposes regulation, and prioritizes individual freedom and commerce.
\item \neither{}: Does not align with convinced or skeptical stance and may present a neutral or unrelated stance.
\end{itemize}

\subsection{Media Frame Annotation}
\label{subsec:anno_media_frame}

Climate change, a critical global issue, refers to long-term alterations in temperature and weather patterns, largely driven by human activities such as fossil fuel combustion. As this issue gains prominence, memes—images paired with text—have become a widespread tool for expressing opinions and social commentary online via media framing.

In this task, you will be given CC memes and will be asked the following question: \textbf{\textit{which media frames are used in these CC memes?}} Choose one or multiple that apply.

\begin{itemize}
\setlength{\itemsep}{0pt} 
\setlength{\parskip}{0pt} 
\setlength{\topsep}{0pt}  
\item  
\real{} emphasizes that there are evidences indicating that CC is occurring; 
\item  
\hoax{} questions the faithfulness of public communication by politicians, the media, environmentalists, etc., e.g., if they are misrepresented or manipulated; 
\item  
\cause{} attributes human activities as a significant cause of CC; 
\item  
\impact{} highlights that CC leads to more net negative outcomes 
 than if there was no CC; 
\item  
\allocation\ captures discussions about the responsibility of certain groups, such as nations, organizations, or even generations, to take action on climate change than others;
\item 
\propriety\ reflects debates on whether current actions are appropriate or effective;
\item  
\adequacy\ highlights critiques regarding whether existing measures are sufficient to address climate risks or more actions are needed;
\item  
\prospect\ explores perceptions of the potential outcomes of positive actions, distinguishing between climate doomists, who view catastrophe as inevitable, and climate risk realists, who believe meaningful prevention is still achievable~\cite{davidson2024climate}.
\end{itemize}

\subsection{Discussion about Moral Frame}
\label{subsec:moral_frame}
While moral judgment is a salient feature in many memes—such as blaming past generations, exposing hypocrisy, or invoking responsibility for future generations—it is theoretically debated whether morality should be treated as a distinct media frame or as an underlying component of framing processes \citep{entman1993framing, gamson1989media}.
Rather than isolating morality as a standalone frame, we integrate moral reasoning into the \allocation\ frame.
This choice reflects our observation that moral claims are often embedded in discussions of responsibility and fairness, and enables us to capture moral stances consistently across both \convinced{} and \skeptical{} memes without losing theoretical coherence.

\subsection{Frame-level IAA} \label{subsec:iaa_appendix}
Table~\ref{tab:iaa} presents  per-frame Cohen’s $\kappa$ for inter-annotator agreement (IAA) among two annotators.

\begin{table}[t]
\centering
\small
\begin{tabular}{ll}
\toprule
Frame & $\alpha$ \\ \midrule
Real     & 0.810 \\
Hoax      & 0.868 \\
Cause     & 0.825 \\
Impact   & 0.711 \\
Action\_allocation & 0.786 \\
Action\_propriety & 0.777 \\
Action\_adequacy & 0.740 \\
Action\_prospect & 0.834 \\
\bottomrule
\end{tabular}
\caption{Cohen's $\kappa$ scores for IAA among two annotators.}
\label{tab:iaa}
\end{table}

\section{Frame Distribution in Climate Change Memes}
\label{sec:frame_distribution}

Among the 1,184 climate change-associated memes we identified, we annotated a total of 2,499 frames, averaging approximately 2.11 frames per meme. Table~\ref{tab:frame_distribution} presents the distribution of these frames across eight main categories. The \adequacy\ frame is the most frequently annotated, followed closely by \hoax, \impact, and \cause frames. \prospect\ is the least frequent frame, indicating fewer memes discussing future-oriented aspects of climate change.

\begin{table}[t]
\centering
\small
\resizebox{0.5\textwidth}{!}{
\begin{tabular}{lcccc}
\toprule
\textbf{Frame} & \textbf{\real} & \textbf{\hoax} & \textbf{\cause} & \textbf{\impact} \\
\midrule
\textbf{Count} & 269 & 387 & 370 & 395  \\
\textbf{\%}    & 10.8\% & 15.5\% & 14.8\% & 15.8\% \\\midrule
\textbf{Frame} & \textbf{\allocation} & \textbf{\propriety} & \textbf{\adequacy} & \textbf{\prospect} \\
\textbf{Count} &
 208 & 382 & 419 & 69 \\
 \textbf{\%}    & 
 8.3\% & 15.3\% & 16.8\% & 2.8\% \\
\bottomrule
\end{tabular}
}
\caption{Distribution of frames annotated in the 1,184 climate change-associated memes (2,499 in total, averaging 2.11 per meme).}
\label{tab:frame_distribution}
\end{table}

\section{Frame Correlation}\label{sec:frame_correlation}

\begin{figure}[t!bh]
    \centering
    \includegraphics[width=0.8\linewidth]{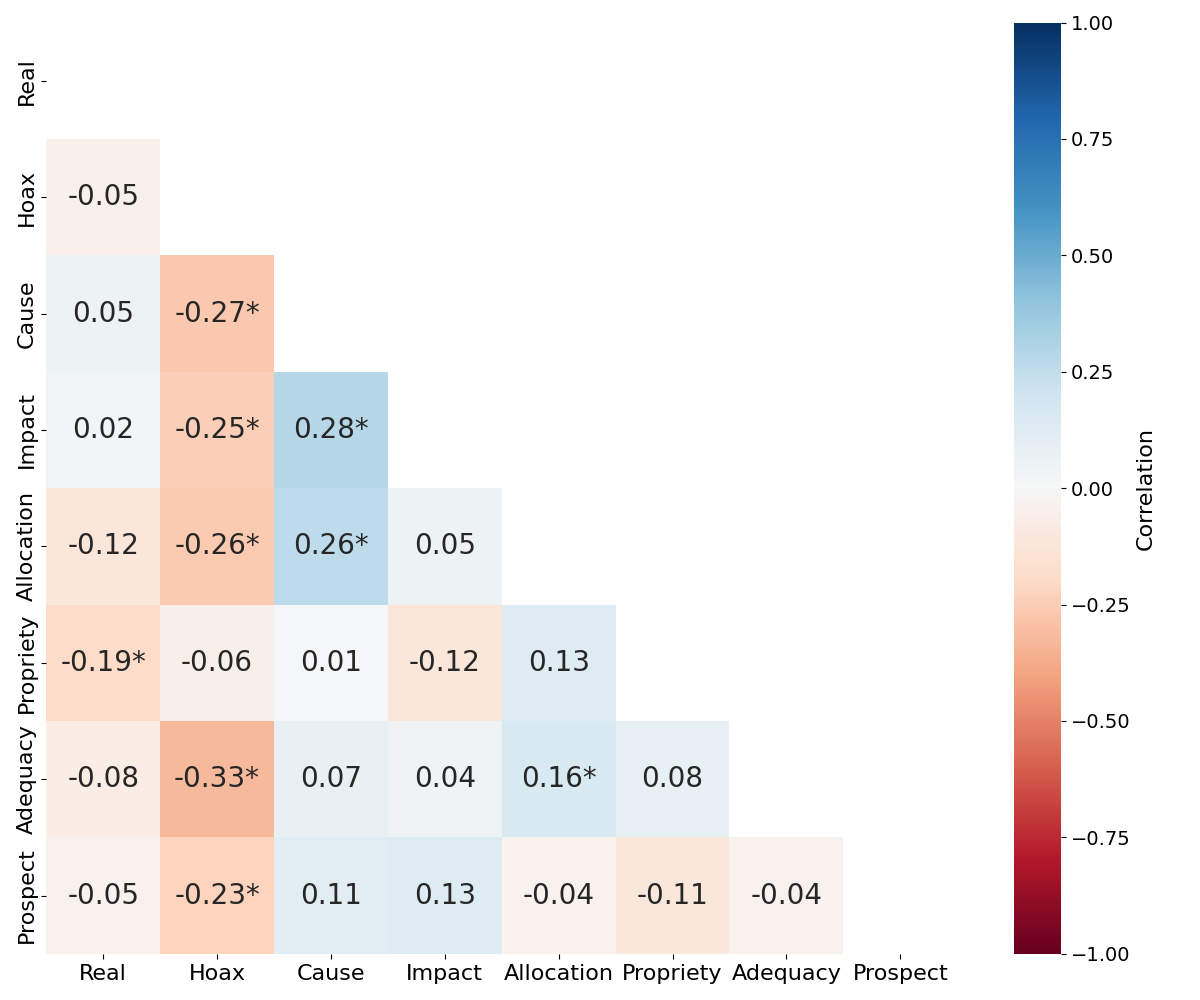}
    \caption{Correlation heatmap of frames. The values represent pairwise Pearson correlation coefficients. Values marked with * indicate corresponding p-values less than 0.05, indicating significance.}
    \label{fig:correlation_heatmap}
\end{figure}

We demonstrate the correlation among 8 frames in Figure~\ref{fig:correlation_heatmap}.

\newpage

\section{Stance and Frame Distribution in Test Set \& Majority Vote Baseline}
\label{sec:majority_vote}

In our dataset, we analyzed a total of 235 cases annotated with stance labels. The distribution of the stances in these cases is summarized in Table~\ref{tab:stance_distribution_appx}. 

\begin{table}[h]
    \centering
    \small
    \begin{tabular}{lccc}
    \toprule
    \textbf{Stance} & \textbf{Convinced} & \textbf{Skeptical} & \textbf{Neither} \\
    \midrule
    \textbf{Count} & 190 & 32 & 13 \\
    \bottomrule
    \end{tabular}
    \caption{Distribution of stance labels in the 235 test memes.}
    \label{tab:stance_distribution_appx}
\end{table}

The majority class is \textit{Convinced} with 190 cases. A majority-vote baseline predicting this class achieves an accuracy of 0.8085 and a macro F1 score of 0.2980.

Table~\ref{tab:frame_distribution_test} summarizes the distribution of ``positive'' and ``negative'' labels across the eight frames.
For each meme, we computed a majority-vote baseline by predicts ``negative'' for all frames.
The reported metrics are averaged over all 235 cases.

\begin{table}[h]
    \centering
    \small
    \begin{tabular}{lrrr}
    \toprule
    \textbf{Frame} & \textbf{Positive} & \textbf{Negative} & \textbf{Total} \\
    \midrule
    \real & 42 & 193 & 235 \\
    \hoax & 79 & 156 & 235 \\
    \cause & 75 & 160 & 235 \\
    \impact & 74 & 161 & 235 \\
    \allocation & 49 & 186 & 235 \\
    \propriety & 78 & 157 & 235 \\
    \adequacy & 82 & 153 & 235 \\
    \prospect & 12 & 222 & 235 \\
    \bottomrule
    \end{tabular}
    \caption{Distribution of ``positive'' and ``negative'' labels across eight frames in the 235 test memes.}
    \label{tab:frame_distribution_test}
\end{table}

This baseline, predicting ``negative'' for all frames, achieves an average accuracy of 0.7383 and an average macro F1 score of 0.4398 across the dataset.

\begin{table}[htbp!]
\centering
\small
\resizebox{\linewidth}{!}{
\begin{tabular}{rllllll}
\toprule

\multicolumn{1}{l}{\textbf{Model}} & \multicolumn{1}{l}{\textbf{\#S}} & \textbf{Inputs} & \textbf{Acc.} & \textbf{F1} & \textbf{precision} & \textbf{recall} \\
\midrule
\multirow{30}{*}{LLaVA} & \multirow{6}{*}{0} & meme           & 76.89 & 28.98 & 26.75 & 31.61 \\
&                    & meme+\ocr{}       & 45.80 & 30.80 & 35.87 & 37.73 \\
&                    & meme+\hum{}     & 68.91 & 46.12 & 58.29 & 55.05 \\
&                    & meme+\hum{}+\ocr{} & 66.39 & 44.83 & 48.50 & 55.75 \\
&                    & meme+\syn{}     & 55.46 & 34.41 & 36.67 & 38.22 \\
&                    & meme+\syn{}+\ocr{} & 56.72 & 35.75 & 38.32 & 40.48 \\
\cline{2-7}
& \multirow{6}{*}{1} & meme           & 64.29 & 37.75 & 39.42 & 38.16 \\
&                    & meme+\ocr{}       & 60.08 & 42.27 & 41.78 & 49.04 \\
&                    & meme+\hum{}     & 73.95 & 55.68 & 54.30 & 63.42 \\
&                    & meme+\hum{}+\ocr{} & 71.85 & 52.60 & 53.35 & 61.04 \\
&                    & meme+\syn{}     & 62.18 & 37.19 & 37.69 & 40.12 \\
&                    & meme+\syn{}+\ocr{} & 66.39 & 41.85 & 41.85 & 44.24 \\
\cline{2-7}
& \multirow{6}{*}{2} & meme           & 68.07 & 34.77 & 40.36 & 35.37 \\
&                    & meme+\ocr{}       & 68.91 & 43.26 & 42.31 & 45.27 \\
&                    & meme+\hum{}     & 82.35 & 56.08 & 55.79 & 58.62 \\
&                    & meme+\hum{}+\ocr{} & 74.37 & 45.31 & 43.33 & 52.29 \\
&                    & meme+\syn{}     & 70.59 & 41.25 & 41.07 & 42.70 \\
&                    & meme+\syn{}+\ocr{} & 73.95 & 43.79 & 44.55 & 44.95 \\
\cline{2-7}
& \multirow{6}{*}{3} & meme           & 73.95 & 35.64 & 45.55 & 35.40 \\
&                    & meme+\ocr{}       & 74.79 & 45.90 & 45.97 & 45.95 \\
&                    & meme+\hum{}     & 83.61 & 54.21 & 54.73 & 55.01 \\
&                    & meme+\hum{}+\ocr{} & 81.09 & 51.65 & 52.37 & 53.97 \\
&                    & meme+\syn{}     & 73.11 & 40.04 & 38.67 & 42.22 \\
&                    & meme+\syn{}+\ocr{} & 76.47 & 41.10 & 39.71 & 42.73 \\
\cline{2-7}
& \multirow{6}{*}{4} & meme           & 77.31 & 39.08 & 49.55 & 40.69 \\
&                    & meme+\ocr{}       & 77.31 & 44.06 & 46.19 & 42.86 \\
&                    & meme+\hum{}     & 86.55 & 56.68 & 62.50 & 54.48 \\
&                    & meme+\hum{}+\ocr{} & 83.19 & 53.57 & 55.13 & 53.97 \\
&                    & meme+\syn{}     & 73.95 & 40.01 & 38.76 & 41.69 \\
&                    & meme+\syn{}+\ocr{} & 76.89 & 41.10 & 39.60 & 42.90 \\
\midrule
\multirow{30}{*}{Molmo} & \multirow{6}{*}{0} & meme           & 42.02 & 25.02 & 30.30 & 30.31 \\
&                    & meme+\ocr{}       & 39.08 & 24.32 & 31.31 & 31.70 \\
&                    & meme+\hum{}     & 47.06 & 33.27 & 40.09 & 46.07 \\
&                    & meme+\hum{}+\ocr{} & 43.70 & 31.50 & 40.16 & 45.55 \\
&                    & meme+\syn{}     & 30.67 & 25.91 & 37.67 & 41.08 \\
&                    & meme+\syn{}+\ocr{} & 28.99 & 24.80 & 37.18 & 40.38 \\
\cline{2-7}
& \multirow{6}{*}{1} & meme           & 61.76 & 30.83 & 30.98 & 32.34 \\
&                    & meme+\ocr{}       & 52.10 & 31.68 & 34.73 & 37.93 \\
&                    & meme+\hum{}     & 59.66 & 41.76 & 43.43 & 51.90 \\
&                    & meme+\hum{}+\ocr{} & 53.78 & 37.00 & 41.04 & 48.83 \\
&                    & meme+\syn{}     & 53.78 & 37.45 & 39.43 & 46.01 \\
&                    & meme+\syn{}+\ocr{} & 56.72 & 39.45 & 41.55 & 48.08 \\
\cline{2-7}
& \multirow{6}{*}{2} & meme           & 45.38 & 26.90 & 31.27 & 32.56 \\
&                    & meme+\ocr{}       & 52.52 & 31.22 & 34.30 & 36.36 \\
&                    & meme+\hum{}     & 68.49 & 47.99 & 46.59 & 55.31 \\
&                    & meme+\hum{}+\ocr{} & 66.39 & 43.66 & 44.16 & 52.27 \\
&                    & meme+\syn{}     & 57.38 & 35.00 & 36.62 & 38.26 \\
&                    & meme+\syn{}+\ocr{} & 63.03 & 36.54 & 36.65 & 41.55 \\
\cline{2-7}
& \multirow{6}{*}{3} & meme           & 47.06 & 28.81 & 33.69 & 37.59 \\
&                    & meme+\ocr{}       & 55.88 & 33.20 & 35.32 & 39.48 \\
&                    & meme+\hum{}     & 68.49 & 47.99 & 47.64 & 53.57 \\
&                    & meme+\hum{}+\ocr{} & 68.49 & 44.41 & 45.90 & 51.40 \\
&                    & meme+\syn{}     & 60.34 & 37.13 & 37.96 & 41.21 \\
&                    & meme+\syn{}+\ocr{} & 63.03 & 35.93 & 36.24 & 40.68 \\
\cline{2-7}
& \multirow{6}{*}{4} & meme           & 47.06 & 28.16 & 32.95 & 34.98 \\
&                    & meme+\ocr{}       & 57.56 & 34.70 & 36.37 & 42.78 \\
&                    & meme+\hum{}     & 72.27 & 49.53 & 50.21 & 55.34 \\
&                    & meme+\hum{}+\ocr{} & 70.17 & 46.52 & 52.67 & 54.70 \\
&                    & meme+\syn{}     & 61.76 & 39.25 & 40.45 & 45.16 \\
&                    & meme+\syn{}+\ocr{} & 65.97 & 38.32 & 37.73 & 43.63 \\
\bottomrule
\end{tabular}
}
\caption{VLMs' performance in terms of accuracy and Macro-F1 on stance detection.
\hum{} = human caption, \syn{} = synthetic caption, \#S = number of shots.}
\label{tab:full_vlm_stance_results}
\end{table}
\begin{table}[htbp!]
\centering
\small
\resizebox{\linewidth}{!}{
\begin{tabular}{rllllll}
\toprule
\multicolumn{1}{l}{\textbf{Model}} & \multicolumn{1}{l}{\textbf{\#S}} & \textbf{Inputs} & \textbf{Acc.} & \textbf{F1} & \textbf{precision} & \textbf{recall} \\
\midrule
\multirow{25}{*}{Mistral}          & \multirow{5}{*}{0}               & \ocr{}        & 46.64         & 33.90       & 37.53              & 42.64           \\
                        &                    & \hum{}     & 66.67 & 51.62 & 50.70 & 65.00 \\
                        &                    & \hum{}+\ocr{} & 62.71 & 47.90 & 47.81 & 60.98 \\
                        &                    & \syn{}     & 39.41 & 29.94 & 37.35 & 40.45 \\
                        &                    & \syn{}+\ocr{} & 41.53 & 30.61 & 36.59 & 41.32 \\
                        \cline{2-7}
                        & \multirow{5}{*}{1} & \ocr{}       & 48.10 & 35.28 & 39.01 & 44.88 \\
                        &                    & \hum{}     & 69.20 & 50.68 & 49.24 & 59.52 \\
                        &                    & \hum{}+\ocr{} & 62.87 & 49.07 & 48.56 & 64.09 \\
                        &                    & \syn{}     & 54.62 & 37.70 & 39.04 & 44.40 \\
                        &                    & \syn{}+\ocr{} & 52.94 & 40.13 & 41.31 & 51.75 \\
                        \cline{2-7}
                        & \multirow{5}{*}{2} & \ocr{}       & 48.74 & 36.19 & 39.70 & 45.89 \\
                        &                    & \hum{}     & 71.73 & 52.68 & 50.88 & 60.56 \\
                        &                    & \hum{}+\ocr{} & 63.45 & 46.47 & 46.22 & 56.28 \\
                        &                    & \syn{}     & 56.78 & 39.92 & 40.15 & 46.82 \\
                        &                    & \syn{}+\ocr{} & 53.81 & 42.10 & 43.23 & 56.23 \\
                        \cline{2-7}
                        & \multirow{5}{*}{3} & \ocr{}       & 52.32 & 38.59 & 40.81 & 49.21 \\
                        &                    & \hum{}     & 77.54 & 57.36 & 55.22 & 62.96 \\
                        &                    & \hum{}+\ocr{} & 69.33 & 50.81 & 49.02 & 59.57 \\
                        &                    & \syn{}     & 55.93 & 36.06 & 37.34 & 41.03 \\
                        &                    & \syn{}+\ocr{} & 57.98 & 43.90 & 43.68 & 55.56 \\
                        \cline{2-7}
                        & \multirow{5}{*}{4} & \ocr{}       & 51.90 & 37.09 & 39.52 & 46.43 \\
                        &                    & \hum{}     & 79.32 & 60.54 & 57.57 & 66.95 \\
                        &                    & \hum{}+\ocr{} & 67.65 & 48.96 & 47.40 & 57.36 \\
                        &                    & \syn{}     & 58.23 & 36.06 & 37.24 & 40.24 \\
                        &                    & \syn{}+\ocr{} & 59.66 & 42.71 & 42.97 & 51.47 \\
                        \midrule
\multirow{25}{*}{Qwen2}            & \multirow{5}{*}{0}               & \ocr{}        & 41.95         & 31.49       & 36.80              & 45.72           \\
                        &                    & \hum{}     & 64.29 & 50.29 & 55.11 & 59.46 \\
                        &                    & \hum{}+\ocr{} & 67.51 & 49.31 & 49.53 & 59.05 \\
                        &                    & \syn{}     & 55.88 & 36.20 & 53.12 & 51.88 \\
                        &                    & \syn{}+\ocr{} & 61.76 & 37.06 & 43.43 & 47.13 \\
                        \cline{2-7}
                        & \multirow{5}{*}{1} & \ocr{}       & 43.28 & 31.34 & 35.86 & 40.83 \\
                        &                    & \hum{}     & 64.71 & 48.60 & 50.08 & 58.11 \\
                        &                    & \hum{}+\ocr{} & 65.13 & 49.02 & 50.31 & 59.80 \\
                        &                    & \syn{}     & 61.34 & 40.18 & 42.22 & 43.47 \\
                        &                    & \syn{}+\ocr{} & 65.13 & 39.49 & 42.33 & 39.37 \\
                        \cline{2-7}
                        & \multirow{5}{*}{2} & \ocr{}       & 44.96 & 32.74 & 37.72 & 44.56 \\
                        &                    & \hum{}     & 70.59 & 52.60 & 55.23 & 59.66 \\
                        &                    & \hum{}+\ocr{} & 69.33 & 50.69 & 52.38 & 61.32 \\
                        &                    & \syn{}     & 62.61 & 40.74 & 50.80 & 43.77 \\
                        &                    & \syn{}+\ocr{} & 64.29 & 42.51 & 47.10 & 46.20 \\
                        \cline{2-7}
                        & \multirow{5}{*}{3} & \ocr{}       & 47.06 & 34.10 & 38.19 & 42.38 \\
                        &                    & \hum{}     & 73.53 & 55.19 & 57.97 & 61.74 \\
                        &                    & \hum{}+\ocr{} & 66.81 & 49.35 & 50.24 & 58.10 \\
                        &                    & \syn{}     & 62.18 & 40.97 & 47.24 & 44.47 \\
                        &                    & \syn{}+\ocr{} & 59.24 & 39.80 & 44.51 & 45.65 \\
                        \cline{2-7}
                        & \multirow{5}{*}{4} & \ocr{}       & 49.16 & 34.06 & 37.96 & 39.98 \\
                        &                    & \hum{}     & 73.11 & 53.28 & 55.79 & 58.30 \\
                        &                    & \hum{}+\ocr{} & 70.17 & 51.66 & 52.22 & 61.01 \\
                        &                    & \syn{}     & 68.91 & 44.66 & 47.97 & 49.62 \\
                        &                    & \syn{}+\ocr{} & 61.34 & 39.08 & 40.29 & 44.12 \\
                        \bottomrule
\end{tabular}
}
\caption{LLMs' performance in terms of accuracy and Macro-F1 on stance detection.
\hum{} = human caption, \syn{} = synthetic caption, \#S = number of shots.}
\label{tab:full_llm_stance_results}
\end{table}
\begin{table}[htbp!]
\centering
\small
\resizebox{\linewidth}{!}{
\begin{tabular}{rllllll}
\toprule
\multicolumn{1}{l}{\textbf{Model}} & \multicolumn{1}{l}{\textbf{\#S}} & \textbf{Inputs} & \textbf{Acc.} & \textbf{F1} & \textbf{precision} & \textbf{recall} \\
\midrule
\multirow{30}{*}{LLaVA} & \multirow{6}{*}{0} & meme           & 46.27 & 41.06 & 44.59 & 50.60 \\
                        &                    & meme+\ocr{}       & 44.65 & 40.27 & 46.50 & 50.39 \\
                        &                    & meme+\hum{}     & 49.38 & 45.18 & 51.77 & 53.19 \\
                        &                    & meme+\hum{}+\ocr{} & 46.70 & 43.10 & 51.23 & 52.08 \\
                        &                    & meme+\syn{}     & 44.93 & 41.34 & 50.59 & 50.22 \\
                        &                    & meme+\syn{}+\ocr{} & 44.19 & 40.85 & 50.65 & 50.35 \\
                        \cline{2-7}
                        & \multirow{6}{*}{1} & meme           & 55.86 & 47.82 & 50.96 & 53.05 \\
                        &                    & meme+\ocr{}       & 53.09 & 45.94 & 49.77 & 52.87 \\
                        &                    & meme+\hum{}     & 56.42 & 49.85 & 53.13 & 56.16 \\
                        &                    & meme+\hum{}+\ocr{} & 56.62 & 49.44 & 53.21 & 55.01 \\
                        &                    & meme+\syn{}     & 57.71 & 49.35 & 52.78 & 53.97 \\
                        &                    & meme+\syn{}+\ocr{} & 56.97 & 48.89 & 52.42 & 53.85 \\
                        \cline{2-7}
                        & \multirow{6}{*}{2} & meme           & 54.87 & 47.03 & 50.80 & 52.67 \\
                        &                    & meme+\ocr{}       & 50.95 & 44.22 & 47.79 & 52.40 \\
                        &                    & meme+\hum{}     & 55.86 & 48.91 & 52.19 & 55.72 \\
                        &                    & meme+\hum{}+\ocr{} & 54.85 & 47.80 & 51.38 & 54.34 \\
                        &                    & meme+\syn{}     & 58.13 & 49.56 & 52.88 & 53.88 \\
                        &                    & meme+\syn{}+\ocr{} & 57.13 & 48.74 & 52.03 & 53.51 \\
                        \cline{2-7}
                        & \multirow{6}{*}{3} & meme           & 53.44 & 46.15 & 49.55 & 53.47 \\
                        &                    & meme+\ocr{}       & 48.86 & 42.50 & 45.73 & 53.41 \\
                        &                    & meme+\hum{}     & 51.91 & 45.14 & 47.76 & 55.38 \\
                        &                    & meme+\hum{}+\ocr{} & 53.56 & 46.75 & 50.18 & 55.09 \\
                        &                    & meme+\syn{}     & 55.62 & 47.85 & 51.38 & 54.45 \\
                        &                    & meme+\syn{}+\ocr{} & 55.53 & 47.74 & 51.29 & 54.54 \\
                        \cline{2-7}
                        & \multirow{6}{*}{4} & meme           & 51.87 & 45.63 & 50.22 & 53.83 \\
                        &                    & meme+\ocr{}       & 46.36 & 40.72 & 43.54 & 53.40 \\
                        &                    & meme+\hum{}     & 49.96 & 44.18 & 47.60 & 55.55 \\
                        &                    & meme+\hum{}+\ocr{} & 50.53 & 44.46 & 48.10 & 54.18 \\
                        &                    & meme+\syn{}     & 52.45 & 45.78 & 49.53 & 54.70 \\
                        &                    & meme+\syn{}+\ocr{} & 52.57 & 45.87 & 49.43 & 54.81 \\
                        \midrule
\multirow{30}{*}{Molmo} & \multirow{6}{*}{0} & meme           & 49.38 & 43.89 & 48.00 & 55.04 \\
                        &                    & meme+\ocr{}       & 53.90 & 46.13 & 49.03 & 55.52 \\
                        &                    & meme+\hum{}     & 53.66 & 47.71 & 51.30 & 56.43 \\
                        &                    & meme+\hum{}+\ocr{} & 53.21 & 47.23 & 50.97 & 56.12 \\
                        &                    & meme+\syn{}     & 51.80 & 45.80 & 49.40 & 55.56 \\
                        &                    & meme+\syn{}+\ocr{} & 52.77 & 46.20 & 49.21 & 55.35 \\
                        \cline{2-7}
                        & \multirow{6}{*}{1} & meme           & 54.88 & 47.73 & 50.06 & 55.65 \\
                        &                    & meme+\ocr{}       & 58.20 & 48.36 & 50.89 & 55.82 \\
                        &                    & meme+\hum{}     & 57.75 & 50.68 & 54.44 & 57.07 \\
                        &                    & meme+\hum{}+\ocr{} & 58.32 & 50.04 & 53.30 & 57.14 \\
                        &                    & meme+\syn{}     & 50.51 & 44.61 & 49.13 & 54.24 \\
                        &                    & meme+\syn{}+\ocr{} & 51.42 & 45.30 & 49.80 & 55.25 \\
                        \cline{2-7}
                        & \multirow{6}{*}{2} & meme           & 53.58 & 46.62 & 48.77 & 54.85 \\
                        &                    & meme+\ocr{}       & 54.89 & 47.35 & 51.36 & 55.12 \\
                        &                    & meme+\hum{}     & 59.84 & 52.15 & 55.45 & 57.88 \\
                        &                    & meme+\hum{}+\ocr{} & 56.88 & 49.58 & 53.43 & 56.63 \\
                        &                    & meme+\syn{}     & 52.73 & 46.66 & 51.48 & 55.47 \\
                        &                    & meme+\syn{}+\ocr{} & 54.26 & 47.58 & 51.83 & 55.43 \\
                        \cline{2-7}
                        & \multirow{6}{*}{3} & meme           & 55.72 & 48.61 & 51.00 & 55.14 \\
                        &                    & meme+\ocr{}       & 54.02 & 47.28 & 51.17 & 56.01 \\
                        &                    & meme+\hum{}     & 60.04 & 52.05 & 54.70 & 57.93 \\
                        &                    & meme+\hum{}+\ocr{} & 58.36 & 50.93 & 54.35 & 57.98 \\
                        &                    & meme+\syn{}     & 56.16 & 49.14 & 53.04 & 55.84 \\
                        &                    & meme+\syn{}+\ocr{} & 53.71 & 47.00 & 50.81 & 55.44 \\
                        \cline{2-7}
                        & \multirow{6}{*}{4} & meme           & 60.37 & 52.60 & 54.99 & 57.41 \\
                        &                    & meme+\ocr{}       & 56.98 & 49.68 & 53.12 & 56.63 \\
                        &                    & meme+\hum{}     & 62.74 & 54.24 & 56.65 & 59.35 \\
                        &                    & meme+\hum{}+\ocr{} & 60.40 & 52.46 & 55.06 & 58.54 \\
                        &                    & meme+\syn{}     & 58.37 & 51.02 & 54.12 & 57.04 \\
                        &                    & meme+\syn{}+\ocr{} & 54.23 & 47.97 & 51.53 & 56.92 \\
 \bottomrule
\end{tabular}
}
\caption{VLMs' performance in terms of accuracy and Macro-F1 on frame detection.
\hum{} = human caption, \syn{} = synthetic caption, \#S = number of shots.}
\label{tab:full_vlm_frame_results}
\end{table}
\begin{table}[htbp!]
\centering
\small
\resizebox{\linewidth}{!}{
\begin{tabular}{rllllll}
\toprule
\multicolumn{1}{l}{\textbf{Model}} & \multicolumn{1}{l}{\textbf{\#S}} & \textbf{Inputs} & \textbf{Acc.} & \textbf{F1} & \textbf{precision} & \textbf{recall} \\
\midrule
\multirow{25}{*}{Mistral} & \multirow{5}{*}{0}               & \ocr{}        & 55.03         & 49.79       & 54.96              & 55.65           \\
                        &                    & \hum{}     & 58.72 & 53.35 & 57.91 & 58.97 \\
                        &                    & \hum{}+\ocr{} & 57.73 & 52.46 & 57.30 & 58.47 \\
                        &                    & \syn{}     & 53.52 & 48.26 & 53.95 & 55.06 \\
                        &                    & \syn{}+\ocr{} & 54.70 & 49.30 & 54.48 & 55.68 \\
                        \cline{2-7}
                        & \multirow{5}{*}{1} & \ocr{}       & 56.32 & 50.75 & 55.43 & 56.04 \\
                        &                    & \hum{}     & 60.81 & 55.08 & 59.07 & 59.85 \\
                        &                    & \hum{}+\ocr{} & 59.79 & 54.18 & 58.18 & 59.29 \\
                        &                    & \syn{}     & 54.97 & 49.76 & 54.58 & 55.17 \\
                        \cline{2-7}
                        &                    & \syn{}+\ocr{} & 56.98 & 51.78 & 56.35 & 57.14 \\
                        & \multirow{5}{*}{2} & \ocr{}       & 59.18 & 52.51 & 55.95 & 56.65 \\
                        &                    & \hum{}     & 62.38 & 56.23 & 59.36 & 60.41 \\
                        &                    & \hum{}+\ocr{} & 62.27 & 55.96 & 58.96 & 60.03 \\
                        &                    & \syn{}     & 56.44 & 50.65 & 55.15 & 55.28 \\
                        &                    & \syn{}+\ocr{} & 58.61 & 52.43 & 56.00 & 56.54 \\
                        \cline{2-7}
                        & \multirow{5}{*}{3} & \ocr{}       & 60.55 & 53.80 & 57.04 & 57.88 \\
                        &                    & \hum{}     & 63.67 & 57.55 & 60.39 & 61.55 \\
                        &                    & \hum{}+\ocr{} & 63.84 & 57.63 & 60.16 & 61.78 \\
                        &                    & \syn{}     & 58.45 & 52.48 & 56.41 & 56.81 \\
                        &                    & \syn{}+\ocr{} & 60.45 & 54.12 & 57.55 & 58.25 \\
                        \cline{2-7}
                        & \multirow{5}{*}{4} & \ocr{}       & 61.71 & 54.79 & 57.59 & 58.68 \\
                        &                    & \hum{}     & 64.61 & 58.31 & 61.00 & 62.31 \\
                        &                    & \hum{}+\ocr{} & 65.09 & 58.78 & 61.06 & 62.94 \\
                        &                    & \syn{}     & 59.03 & 53.01 & 56.42 & 57.48 \\
                        &                    & \syn{}+\ocr{} & 61.78 & 55.20 & 58.10 & 59.12 \\
                        \midrule
\multirow{25}{*}{Qwen2}   & \multirow{5}{*}{0}               & \ocr{}        & 56.60         & 49.47       & 53.29              & 54.49           \\
                        &                    & \hum{}     & 65.07 & 55.32 & 57.85 & 58.63 \\
                        &                    & \hum{}+\ocr{} & 60.82 & 52.94 & 55.81 & 56.99 \\
                        &                    & \syn{}     & 54.24 & 47.82 & 52.28 & 53.33 \\
                        &                    & \syn{}+\ocr{} & 53.69 & 47.41 & 51.66 & 53.06 \\
                        \cline{2-7}
                        & \multirow{5}{*}{1} & \ocr{}       & 59.59 & 51.82 & 54.94 & 55.63 \\
                        &                    & \hum{}     & 64.07 & 56.54 & 59.53 & 60.69 \\
                        &                    & \hum{}+\ocr{} & 63.39 & 55.68 & 58.50 & 59.15 \\
                        &                    & \syn{}     & 57.55 & 50.92 & 54.79 & 54.84 \\
                        &                    & \syn{}+\ocr{} & 57.69 & 51.08 & 54.90 & 55.30 \\
                        \cline{2-7}
                        & \multirow{5}{*}{2} & \ocr{}       & 62.11 & 54.01 & 56.95 & 57.72 \\
                        &                    & \hum{}     & 64.84 & 57.34 & 59.96 & 61.24 \\
                        &                    & \hum{}+\ocr{} & 64.31 & 56.57 & 58.88 & 60.05 \\
                        &                    & \syn{}     & 58.36 & 52.09 & 55.99 & 57.10 \\
                        &                    & \syn{}+\ocr{} & 60.65 & 53.87 & 57.07 & 58.42 \\
                        \cline{2-7}
                        & \multirow{5}{*}{3} & \ocr{}       & 63.14 & 54.32 & 56.92 & 57.65 \\
                        &                    & \hum{}     & 65.76 & 58.22 & 60.59 & 61.94 \\
                        &                    & \hum{}+\ocr{} & 65.32 & 57.52 & 59.76 & 61.05 \\
                        &                    & \syn{}     & 59.85 & 53.24 & 56.58 & 57.65 \\
                        &                    & \syn{}+\ocr{} & 60.85 & 54.16 & 57.57 & 58.96 \\
                        \cline{2-7}
                        & \multirow{5}{*}{4} & \ocr{}       & 64.02 & 55.45 & 57.99 & 58.80 \\
                        &                    & \hum{}     & 65.86 & 58.23 & 60.66 & 62.18 \\
                        &                    & \hum{}+\ocr{} & 64.98 & 57.51 & 60.01 & 61.12 \\
                        &                    & \syn{}     & 60.33 & 53.98 & 57.27 & 58.26 \\
                        &                    & \syn{}+\ocr{} & 60.88 & 54.24 & 57.45 & 58.45 \\
 \bottomrule
\end{tabular}
}
\caption{LLMs' performance in terms of accuracy and Macro-F1 on frame detection.
 \hum{} = human caption, \syn{} = synthetic caption, \#S = number of shots.}
\label{tab:full_llm_frame_results}
\end{table}

\section{Additional Experimental Results}  \label{sec:full_results}

\begin{figure*}[htbp!]
    \centering
    \begin{subfigure}[b]{0.98\linewidth}
        \centering
        \includegraphics[width=0.98\linewidth]{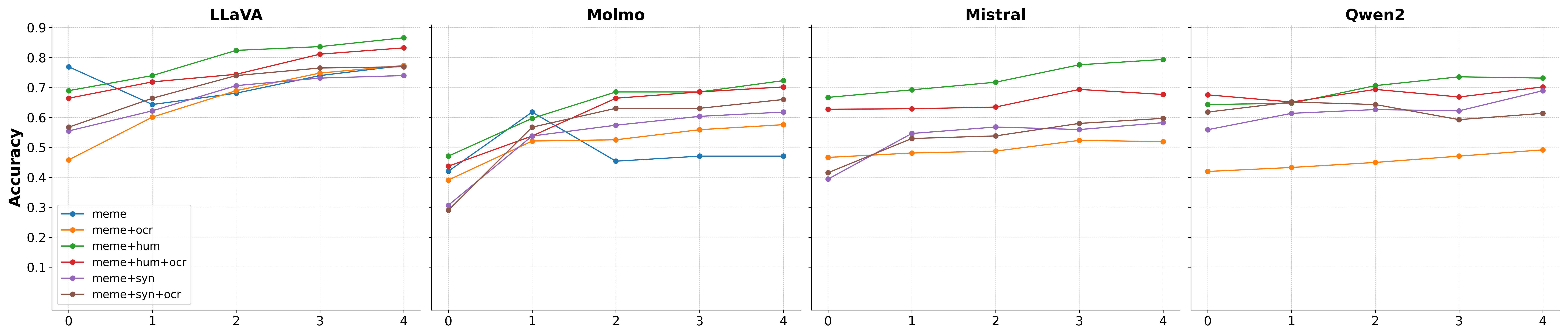} 
    \end{subfigure}
    \hfill 
    \begin{subfigure}[b]{0.98\linewidth}
        \centering
        \includegraphics[width=0.98\linewidth]{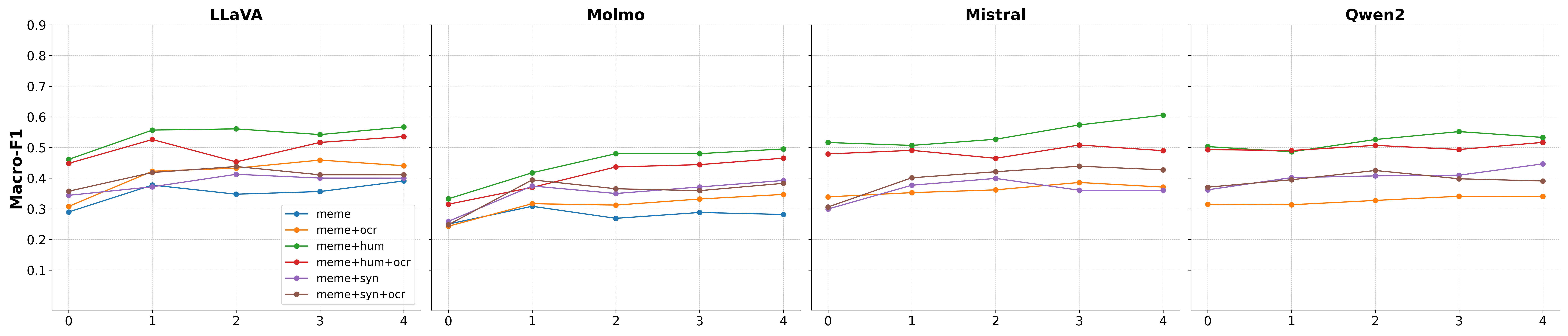}
        \caption{Stance detection}
    \end{subfigure}
    \hfill 
    \begin{subfigure}[b]{0.98\linewidth}
        \centering
        \includegraphics[width=0.98\linewidth]{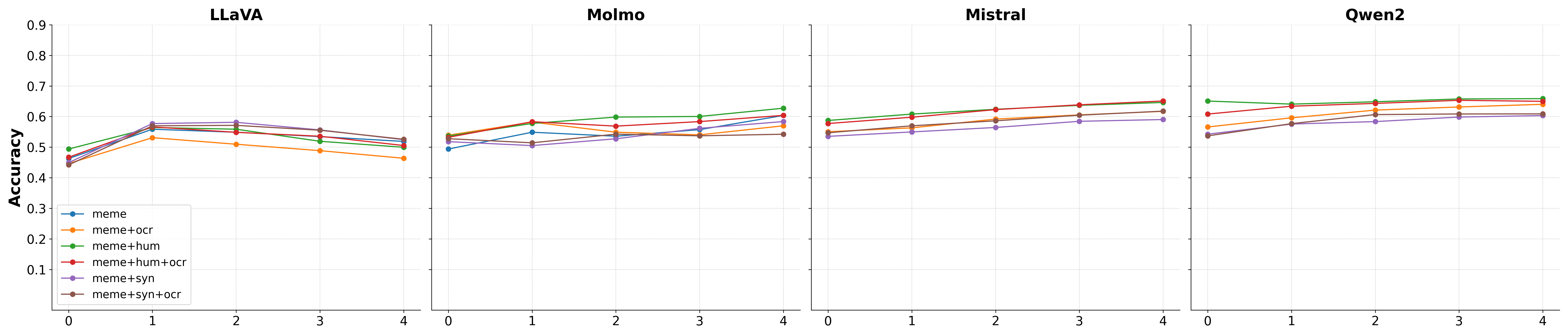}
    \end{subfigure}
    \hfill 
    \begin{subfigure}[b]{0.98\linewidth}
        \centering
        \includegraphics[width=0.98\linewidth]{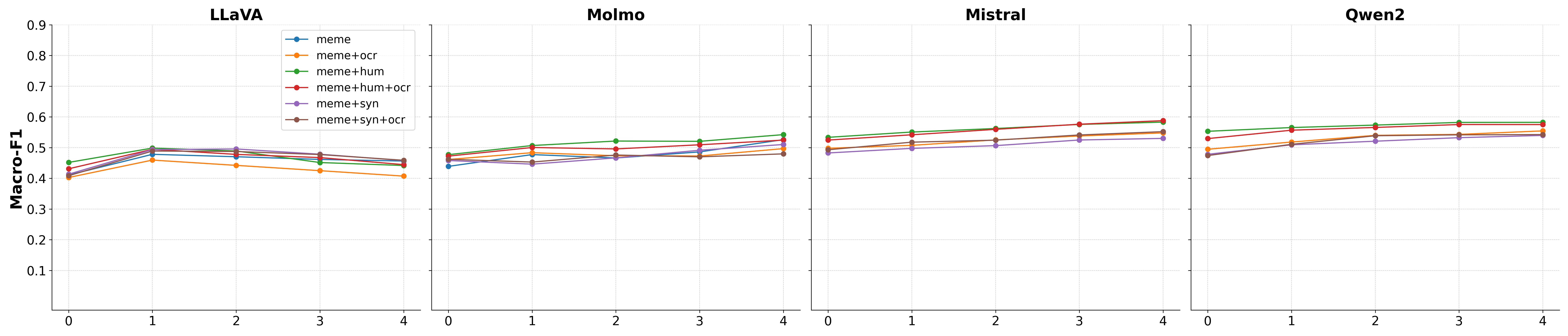}
        \caption{Frame detection}
    \end{subfigure}
    \caption{Accuracy and Macro-F1 of VLMs on stance and media frame detection with different shot and input setups.
    }
    \label{fig:ablation_plots}
\end{figure*}

We show the full experimental results of stance detection of VLMs in Table~\ref{tab:full_vlm_stance_results}, of LLMs in Table~\ref{tab:full_llm_stance_results}, frame detection of VLMs in Table~\ref{tab:full_vlm_frame_results}, and of LLMs in Table~\ref{tab:full_llm_frame_results}.
In line chart Figure~\ref{fig:ablation_plots}, we also present VLM performances on stance and media frame detection with different shot and input setups.

\newpage

\section{Definition of CC-associated communication science concepts}  \label{sec:cs_concepts}

\paragraph{Humor Type.}
For the humor types category, the content format used to create humor in memes is coded. Humor types are initially independent of the respective humor style. Following~\citet{taecharungroj2015humour}, a distinction is made between seven humor types, several of which can in principle be used simultaneously in a meme.

    \begin{itemize}
    \setlength{\itemsep}{0pt} 
    \setlength{\parskip}{0pt} 
    \setlength{\topsep}{0pt}  
    \item  
    \textbf{Puns} use language to construct new meanings or use words or phrases in a way that suggests two interpreta-tions, e.g. words that are pronounced the same but have different meanings. 
    \item  
    \textbf{Personifications} (personification) are used when human s and/or behavior are attributed to other objects such as animals, plants or ob-jects. 
    \item  
    \textbf{Exaggerations} and understatements are disproportionate enlargements or reductions of a fact or context. Something is depicted as being larger or smaller than it (supposedly) actually is. Both the behavior of people and the consequences of events are depicted larger or smaller.
    \item  
    \textbf{Comparisons} are combinations of two or more elements (e.g., before and after pictures) to construct a funny situation. 
    \item  
    \textbf{Irony} and \textbf{sarcasm} refer to the use of words to express the opposite of what one actually means. 
    \item 
    \textbf{Surprise} is the use of unexpected elements in memes. Memes with this element have a surprising ending/resolution. 
    \item  
    \textbf{Jokes} and nonsense describes content with no particular meaning and non-serious statements or actions that are only in-tended to make us laugh. 
    \end{itemize}

\paragraph{Personalization.} Who is shown in the picture?

    \begin{itemize}
    \setlength{\itemsep}{0pt} 
    \setlength{\parskip}{0pt} 
    \setlength{\topsep}{0pt}  
    \item
    \textbf{Political} actors include heads of state, members of government, official state delegates to the COP, ministers, representatives of institutions such as the UN or EU. 
    \item  
    \textbf{NGO} members or environmental activists. Members of protest movements such as Fridays for Future are considered environ-mental activists, whereas ``normal'' participants in demonstrations are coded as ``normal citizens''. 
    \item  
    \textbf{Celebrities} are famous people who do not have an official political function. This includes, for example, people such as athletes, actors, influencers or artists.
    \item  
    \textbf{Normal} or ordinary citizens are people who are not clearly assigned to one of the other categories.
    \end{itemize}

\paragraph{Responsibility.} To whom the responsibility for solving or combating the climate problem is attributed. The aim is to record who should take measures against climate change (e.g. more environmental protection, fewer emissions) or who isexpected to do so. Responsibility can be explicitly attributed or suggested by listing necessary measures that only a certain group can take.

    \begin{itemize}
    \setlength{\itemsep}{0pt} 
    \setlength{\parskip}{0pt} 
    \setlength{\topsep}{0pt}  
    \item
    Responsibility at \textbf{micro} level: Responsibility for individual persons such as politicians, activists, entrepreneurs, etc.
    \item  
    Responsibility at \textbf{meso} level: Responsibility for individual companies, institutions, parties, parliaments, governments.
    \item  
    Responsibility at \textbf{macro} level: Responsibility for certain countries, politics in general, the economic system, society, us as humanity, etc
    \end{itemize}

\section{Case Study: Memes with Features in Communication}
\label{sec:error_analysis}

We present a case study on selected memes to analyze the model’s stance and frame prediction performance.
Table~\ref{tab:qualitative_examples} showcases three representative examples where our model made prediction errors. For each meme, we provide the associated features, gold and predicted stances and frames, as well as a detailed interpretation generated by the model.
This qualitative analysis helps illustrate common challenges and nuanced aspects of meme communication that contribute to model misclassification.

\begin{table*}[ht]
\centering
\small
\resizebox{0.98\textwidth}{!}{
\begin{tabular}{m{0.12\textwidth}ccc}
\toprule
& \textbf{Meme 1} & \textbf{Meme 2} & \textbf{Meme 3} \\
& \includegraphics[width=0.28\textwidth]{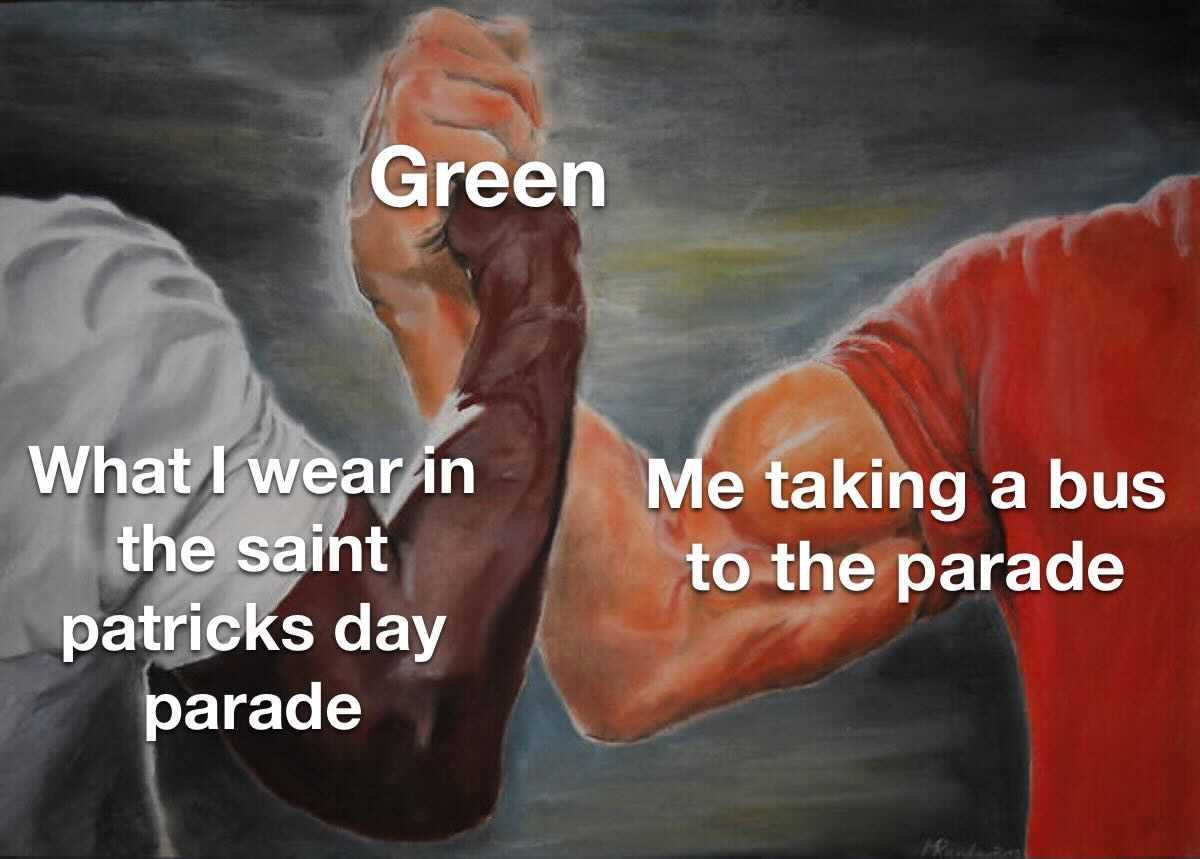}
& \includegraphics[width=0.28\textwidth]{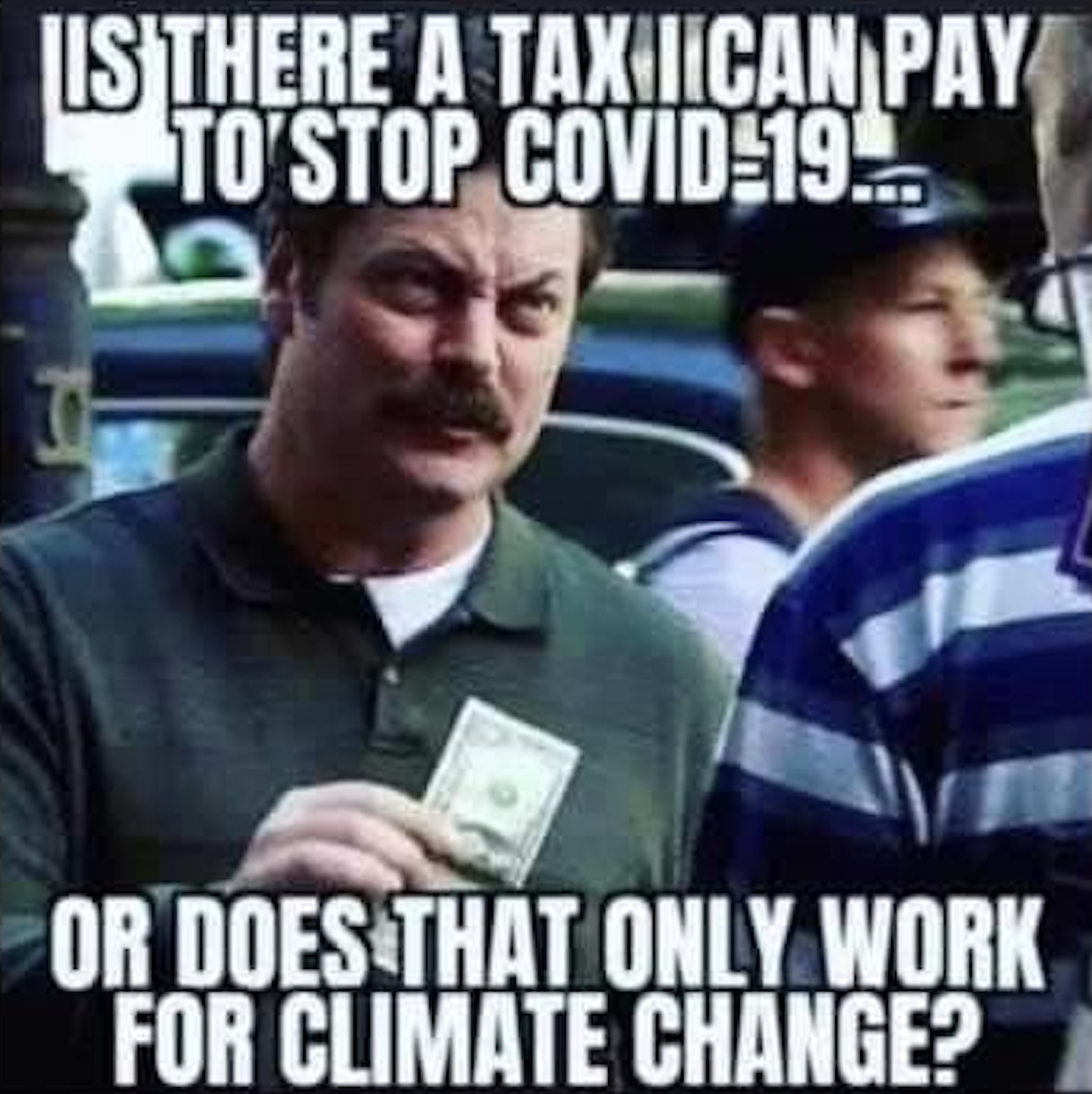}
& \includegraphics[width=0.28\textwidth]{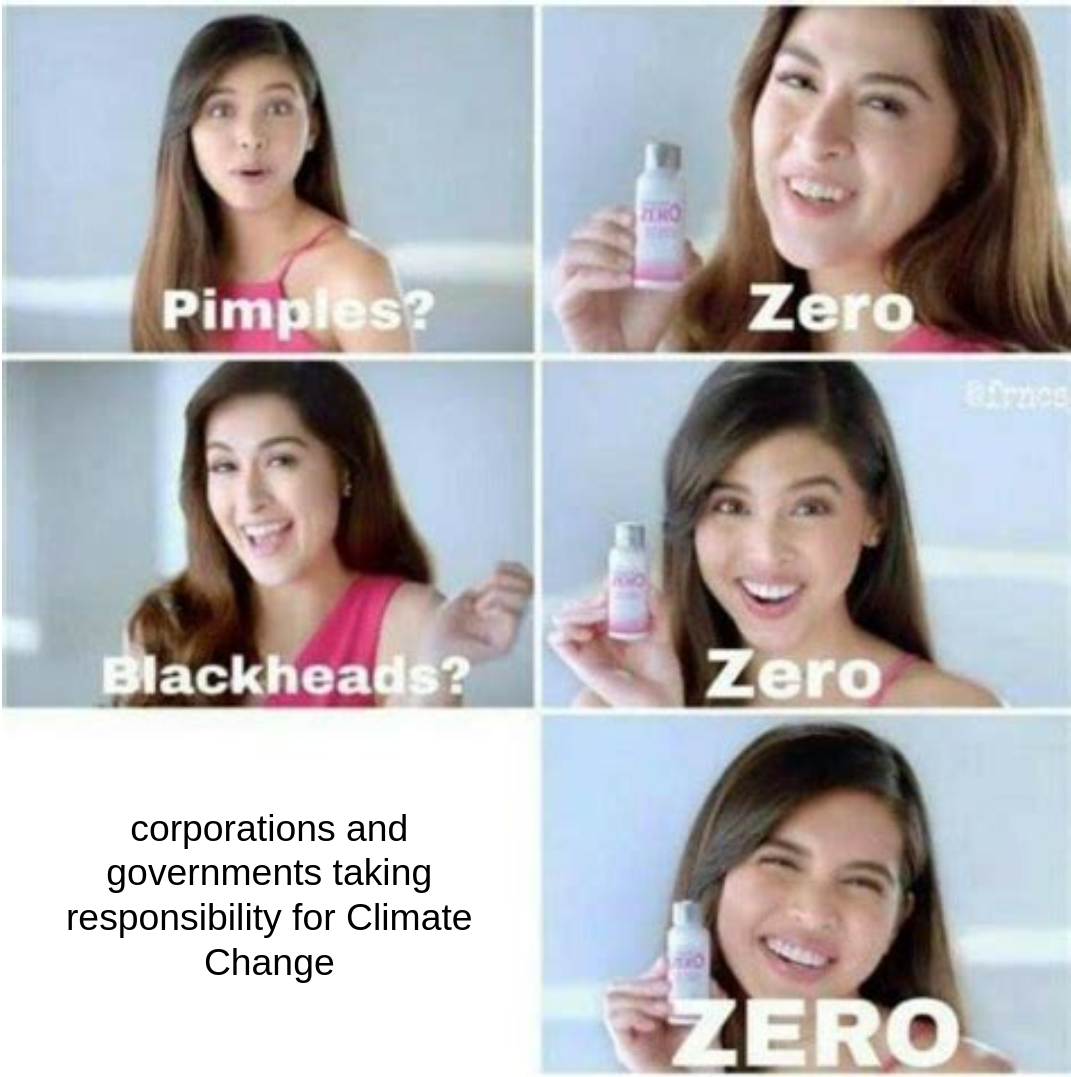} \\
\midrule
\textbf{Features} 
& \texttt{responsib\_micro}, \texttt{humortypes\_pun}
& \texttt{responsib\_macro}
& \texttt{humortypes\_irony} \\
\midrule
\textbf{Gold Stance} 
& Convinced 
& Skeptical 
& Convinced \\
\midrule
\textbf{Predicted Stance} 
& Neither 
& Convinced 
& Skeptical \\
\midrule
\textbf{Gold Frames} 
& \textsc{Allocation}, \textsc{Prospect} 
& \textsc{Adequacy} 
& \textsc{Hoax}, \textsc{Propriety}, \textsc{Adequacy} \\
\midrule
\textbf{Predicted Frames} 
& \textsc{Real}, \textsc{Impact} 
& \textsc{Real}, \textsc{Hoax}, \textsc{Adequacy} 
& \textsc{Real}, \textsc{Impact} \\
\midrule
\textbf{Model Interpretation} 
& \begin{minipage}[t]{0.3\textwidth}
This meme humorously highlights the common experience of getting unexpected looks from strangers during St. Patrick's Day celebrations. The image cleverly combines two related scenarios: wearing green for the holiday and taking public transportation. By showing muscular arms representing these two situations joining together, it suggests that both activities share a common outcome - drawing attention or being stared at. The meme taps into the relatable discomfort of standing out in a crowd, particularly during a holiday known for its vibrant green attire and large gatherings. Its simplicity and relatability make it an effective piece of internet humor that resonates with many people's St. Patrick's Day experiences.
\end{minipage}
& \begin{minipage}[t]{0.3\textwidth}
The meme features a man holding a dollar bill, with a caption that reads, "Is there a tax I can pay to stop COVID-19... Or does that only work for climate change?" This meme is a humorous take on the perceived difference in public opinion and government response to the COVID-19 pandemic and climate change. The implication is that people are more willing to pay taxes to address climate change, but not to combat the COVID-19 pandemic. The meme plays on the idea that some individuals prioritize environmental issues over public health crises, and it uses the man's expression to convey a sense of frustration or disbelief at this perceived attitude. The humor arises from the juxtaposition of the man's action (offering to pay a tax) with the implication that such an action would be more effective for climate change than for COVID-19.
\end{minipage}
& \begin{minipage}[t]{0.3\textwidth}
The meme features a series of images of a woman with a surprised or shocked expression, holding a bottle of ''ZERO`` with the text ''Pimples? Zero`` and ''Blackheads? Zero`` above each image. The humor in the meme comes from the exaggerated and unexpected response to the common skin issues of pimples and blackheads. The woman's expression and the repeated use of the word ''zero`` suggest that the product being advertised is highly effective in eliminating these skin problems. The meme plays on the idea that the product is so powerful that it can completely eradicate these issues, which is an exaggerated and humorous claim. The meme is likely intended to be a parody of skincare product advertising, using hyperbole and visual cues to create a comedic effect.
\end{minipage} \\
\bottomrule
\end{tabular}
}
\caption{Qualitative examples of meme stance and frame prediction with model errors and interpretation generated by LLaVA.}
\label{tab:qualitative_examples}
\end{table*}

\section{Elaboration About Use of AI Assistants}
The authors used ChatGPT to polish writings for clarity and coherence and to assist with code generation.
The authors manually inspected all ChatGPT suggestions and made corrections when necessary.

\end{document}